\documentclass[journal]{IEEEtran}
\usepackage{amsmath,epsfig,color}
\usepackage{url}
\usepackage{bigstrut,multirow,rotating}
\usepackage{textcomp,booktabs}
\usepackage{tabularx}
\usepackage{floatrow}
\newfloatcommand{capbtabbox}{table}[][\FBwidth]
\usepackage{colortbl}
\usepackage{xcolor}
\usepackage{multirow}
\usepackage{booktabs}
\usepackage{underscore}
\usepackage[marginal]{footmisc}
\usepackage{subfigure}
\usepackage{algorithm,algorithmic}
\usepackage{amssymb}
\usepackage{graphicx}
\usepackage{bbding}
\usepackage{makecell}
\usepackage{float}
\usepackage{threeparttable}
\usepackage{cite}
\usepackage[linkcolor=black]{hyperref}
\begin{document}
\newcommand{\bl}[1]{{\color{black}#1}}
\newcommand{\br}[1]{{\color{red}#1}}
\newcommand{\bbl}[1]{{\color{black}#1}}
\newcommand{\eg}{\textit{e}.\textit{g}.}
\title{Deep Feature Statistics Mapping for Generalized Screen Content Image Quality Assessment}
\author{Baoliang Chen, Hanwei Zhu, Lingyu Zhu, Shiqi Wang,~\IEEEmembership{Senior Member,~IEEE} and Sam Kwong,~\IEEEmembership{Fellow,~IEEE}
\thanks{This work is supported in part by the Shenzhen Science and Technology Program under Project JCYJ20220530140816037; in part by the Hong Kong Innovation and Technology Commission [InnoHK Project Centre for Intelligent Multidimensional Data Analysis (CIMDA)]; in part by the Hong Kong Research Grants Council (RGC) of the General Research Fund (GRF) under Grant 11203220 (CityU 9042957); and in part by the ITF GHP/044/21SZ.}

\thanks{Baoliang Chen is with the Department of Computer Science, South China Normal University, China (e-mail: blchen6-c@my.cityu.edu.hk).}

\thanks{Hanwei Zhu and Lingyu Zhu are with the Department of Computer Science, City University of Hong Kong, Hong Kong (e-mail: hwzhu4-c@my.cityu.edu.hk and lingyzhu-c@my.cityu.edu.hk).} 

\thanks{Shiqi Wang is with the Department of Computer Science, City University of Hong Kong, Hong Kong, China, and also with the City University of Hong Kong, Shenzhen Research Institute, Shenzhen 518057, China (e-mail: shiqwang@cityu.edu.hk).}

\thanks{Sam Kwong is with Lingnan University, Hong Kong (e-mail: samkwong@ln.edu.hk).}

\thanks{Corresponding author: Shiqi Wang.}

}
\maketitle

\begin{abstract}
The statistical regularities of natural images, referred to as natural scene statistics, play an important role in no-reference image quality assessment. However, it has been widely acknowledged that screen content images (SCIs), which are typically computer generated, do not hold such statistics. Here we make the first attempt to learn the statistics of SCIs, based upon which the quality of SCIs can be effectively determined. The underlying mechanism of the proposed approach is based upon the mild assumption that the SCIs, which are not physically acquired, still obey certain statistics that could be understood in a learning fashion. We empirically show that the statistics deviation could be effectively leveraged in quality assessment, and the proposed method is superior when evaluated in different settings. Extensive experimental results demonstrate the Deep Feature Statistics based SCI Quality Assessment (DFSS-IQA)  model delivers promising performance compared with existing NR-IQA models and shows a high generalization capability in the cross-dataset settings.  The implementation of our method is publicly available at
\url{https://github.com/Baoliang93/DFSS-IQA}.
\end{abstract}

\IEEEpeerreviewmaketitle

\section{Introduction}

\IEEEPARstart{S}{creen}  content images (SCIs) have attracted dramatic attention in recent years with the rapid development of digital devices and online communication~\cite{wang2016just,wang2016reduced,wang2015joint}. The SCI appears in various media, such as electronic books, online news, and screen sharing. 
Serving as the fundamental technology in SCI compression, exhibition, and processing,  assessing the quality of SCI automatically becomes a highly demanding task. In practice, due to the imperfect transmission condition and limited storage space, the quality of SCIs can be degraded by several distortion types. 

To assess the image quality, the most reliable way is to collect subjective opinions by crowdsourcing \cite{xu2016parsimonious,xu2013online,xu2013robust,xu2012online,xu2012hodgerank}. However, the subjective study is usually laborious and expensive.
To count for this, numerous objective methods have been proposed for SCI quality assessment (SCIQA)~\cite{min2021screen}.
According to the availability of reference images, the SCIQA methods fall into two categories: full-reference (FR) SCIQA and no-reference (NR) SCIQA. For FR-SCIQA, the hand-crafted features such as structural features, luminance features, and Gabor features are exploited in \cite{fang2017objective, ni2017esim, ni2018gabor}. With the popularisation of deep learning, the convolutional neural network (CNN) based models have been widely investigated in  \cite{zhang2018quality,  yang2021full}. For example, 
Yang \textit{et al.} adopted a fully convolutional network to separate the image into structural regions and texture regions and extracted the structural features and perceptual features from the two regions for SCIQA \cite{yang2021full}.
Compared with the FR-SCIQA, the NR-SCIQA is more practical in real-world applications. In \cite{gu2017no, fang2017no}, the quality-aware features including the picture complexity, brightness, sharpness, and 
textures are adopted. Driven by the hypothesis that the human visual system (HVS) is highly sensitive to sharp edges, Zheng \textit{et al.} divided the SCI into sharp edge regions and non-sharp edge regions \cite{zheng2019no}. Recently, the deep-learning based NR-SCIQA models have shown superior performance by learning the quality-aware features from the data \cite{jiang2019deep, cheng2018fast, chen2018naturalization,chen2021no,yang2021staged}.  In particular, Chen \textit{et al.} incorporated a naturalization module in stacked CNNs for quality-aware feature learning~\cite{chen2018naturalization}. Yang \textit{et al.} proposed a multi-task learning framework which both the distortion types and distortion degrees were analyzed in \cite{yang2021staged}.   However, despite the success of those methods, they usually suffered from the over-fitting problem, presenting an inferior generalization capability, especially in cross-dataset settings. 

\begin{figure}[t]
\begin{minipage}[b]{0.9\linewidth}
 \centerline{\includegraphics[width=1\linewidth]{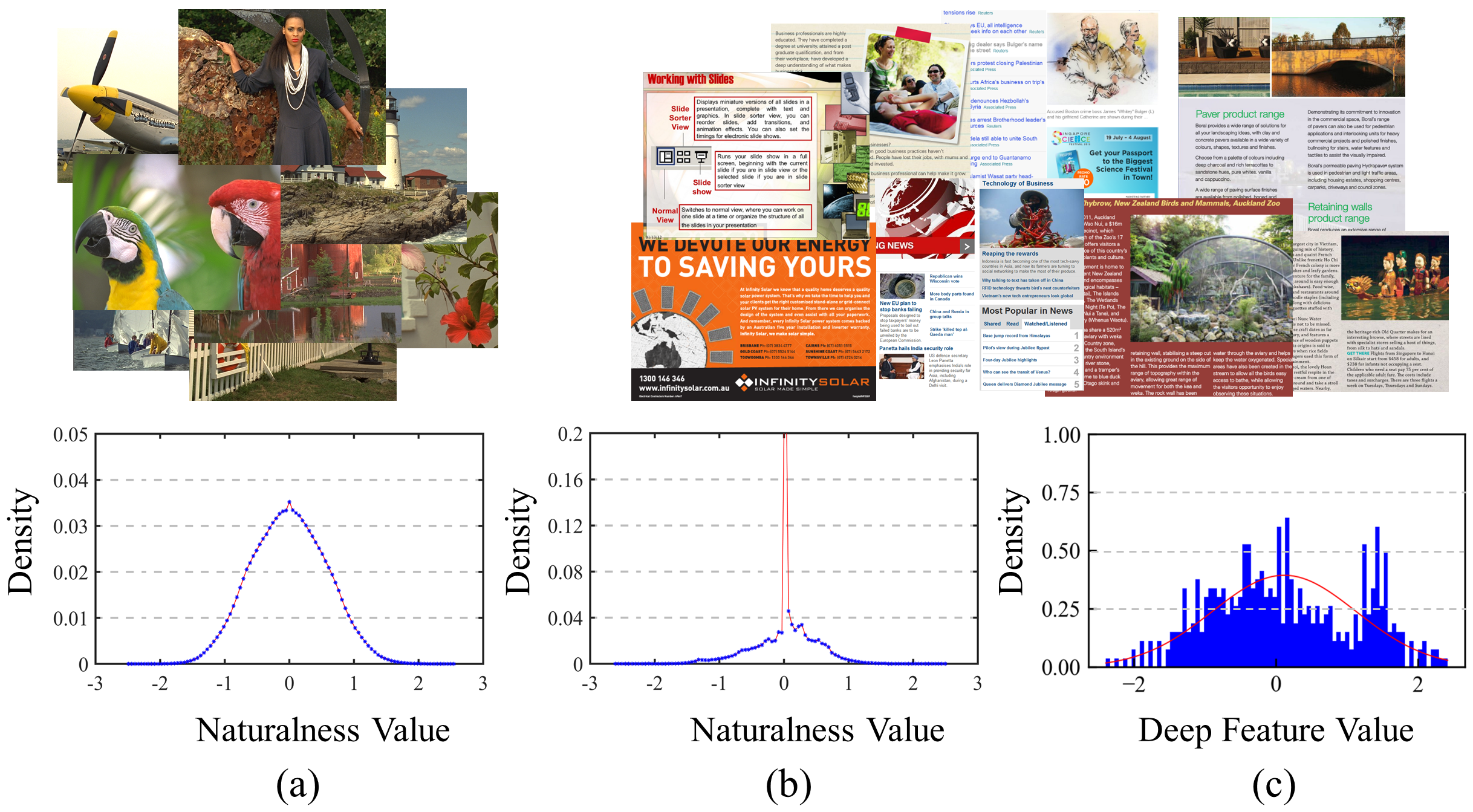}}
\end{minipage}
\caption{\bbl{Illustrations of the statistics of NIs and SCIs. (a) Distribution of Naturalness Values (DNV) of reference images in TID2013 dataset~\cite{ponomarenko2015image}; (b)  DNV of  reference images in SIQAD dataset~\cite{yang2015perceptual}; (c) The deep feature statistics of reference images in SIQAD dataset obtained by our proposed method.}}
\label{nss}
\end{figure}

We believe that the statistics-based methods, which have been proven to be effective for the quality assessment of natural images (NIs), become pivotal for SCIs as well. Typically, natural scene statistics  (NSS) which have been explored with the assumption that natural images usually hold specific statistical properties and such properties will be corrupted when the images are distorted~\cite{mittal2012making, mittal2012no, moorthy2010two,moorthy2011blind}. As a result, the image quality can be promisingly evaluated by measuring the corruption level.  
Despite the impressive effectiveness of NSS for NIs, unfortunately, the NSS may not be statistically meaningful for SCIs.
As shown in Fig.~\ref{nss}, the distributions of naturalness values  \cite{mittal2012no} of  NIs and SCIs present a stark difference, due to the fact that SCIs are computer generated instead of being acquired by optical cameras. As a consequence, the IQA models designed for NIs usually show a dramatic performance degradation on SCIs \cite{gu2016saliency,chen2021no}.

In this paper, we focus on exploring the specific statistics of SCIs in the deep feature domain and developing a generalized NR-SCIQA model. In general, the expected feature statistics should be {shared by different image content} and be aware of the quality degradation. In this regard, a unified distribution regularization is imposed on the feature space, and the relationship between the image quality and the feature statistics is established by regressing the image quality from the distribution destruction levels.
To verify the performance of our method, we conduct both intra-dataset and cross-dataset experiments on two widely used databases, including SIQAD \cite{yang2015perceptual}  and SCID~\cite{ni2017scid} databases. Experimental results have demonstrated the superior performance of our method in terms of both intra-dataset prediction accuracy and cross-dataset generalization capability. The main contributions of our paper are summarized as follows,
\bl{
\begin{itemize}
\item We make the first attempt to address the SCIQA from the perspective of learning SCI statistics. In particular, we learn the statistics of SCIs in the deep-feature domain, then the quality of the distorted image can be estimated by the destruction level of the statistics.

\item  We design a Maximum Mean Discrepancy (MMD) based regularization term to align the distributions of pristine SCIs and normalized distributions of distorted SCIs, in which way the distribution deviation caused by the distortions can be efficiently measured by Kullback-Leibler (KL) divergence.

\item  We propose to disentangle the quality perception into image semantics extraction and distortion acquisition. In particular, an attention mechanism is designed to synthesize the image semantics and distortion to mimic the process of human quality rating.
\end{itemize}
}

\begin{figure*}[t]
\begin{minipage}[b]{1.0\linewidth}
  \centerline{\includegraphics[width=1\linewidth]{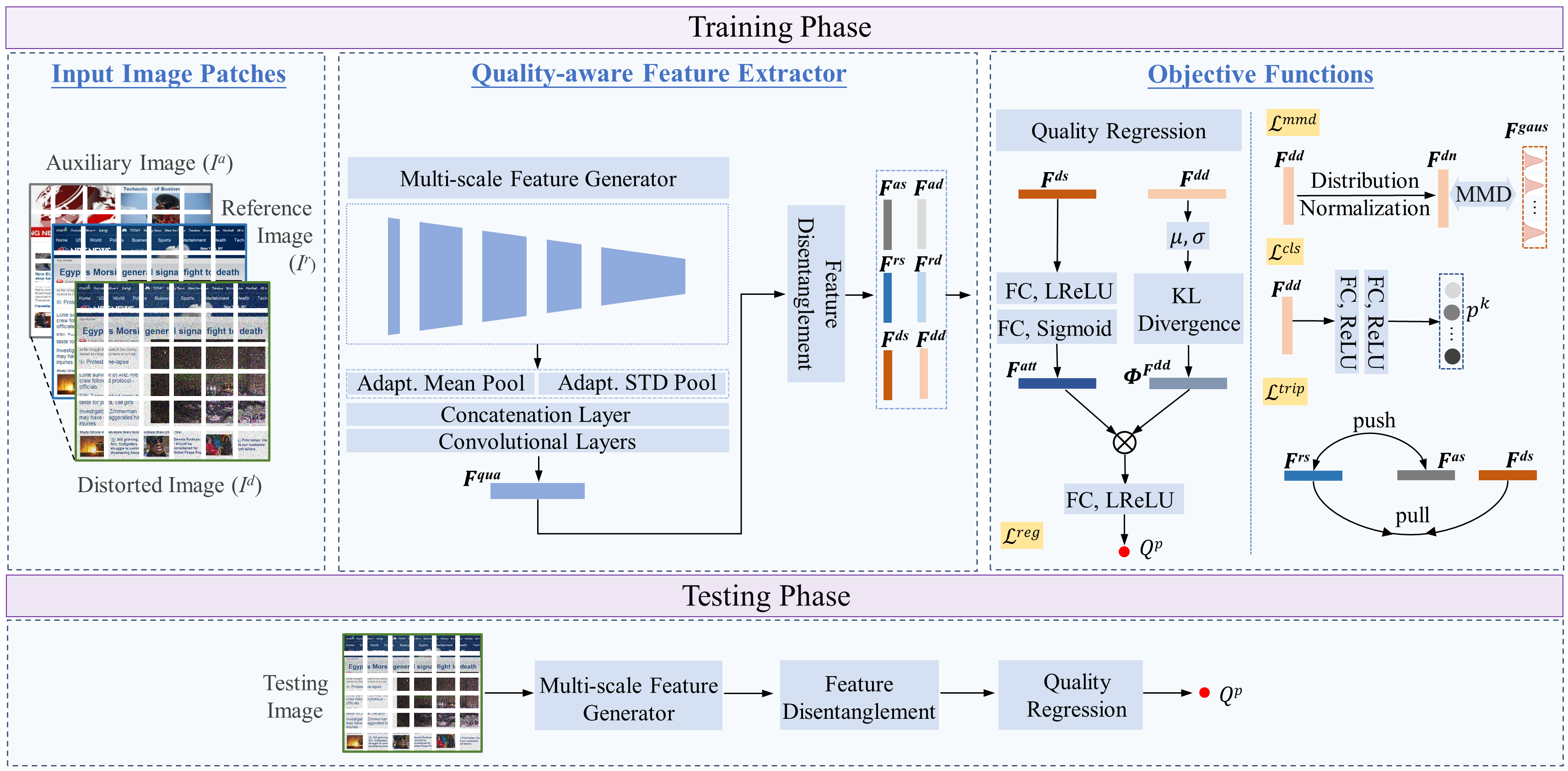}}
\end{minipage}
\caption{{\bbl{Illustration of the framework of our proposed method.  In the training phase, the images are grouped with a triplet including a reference image ($\boldsymbol{I^{r}}$), a distorted image ($\boldsymbol{I^{d}}$), and an auxiliary image ($\boldsymbol{I^{a}}$). In particular, $\boldsymbol{I^{d}}$ shares the same content with $\boldsymbol{I^{r}}$ while its quality is degraded by the distortion.  $\boldsymbol{I^{a}}$ is sampled from pristine images but its content is different from $\boldsymbol{I^{r}}$. Then the quality-aware feature of each image is extracted via a multi-scale feature generator and further disentangled into a semantic-aware feature ($\boldsymbol{F^{rs}}$, $\boldsymbol{F^{ds}}$, $\boldsymbol{F^{as}}$) and a distortion-aware feature ($\boldsymbol{F^{rd}}$, $\boldsymbol{F^{dd}}$, $\boldsymbol{F^{ad}}$). We force the normalized distortion-aware feature to obey a unified distribution ($\boldsymbol{F^{gaus}}$) and treat the unified distribution as the feature  statistics shared by the SCIs. As a consequence, the distortion of the $\boldsymbol{I^{d}}$ can be measured by the feature distribution divergence estimation. Finally, the quality of $\boldsymbol{I^{d}}$ can be regressed by incorporating both its semantic information ($\boldsymbol{F^{ds}}$) and  distortion information ($\boldsymbol{F^{dd}}$). In the testing phase, only the testing image (without reference) is needed for quality prediction.} }}
\label{fig:frame}
\end{figure*}
\section{Related Works}
\bl{
\subsection{No-reference Image Quality Assessment Methods}
Getting rid of the requirement of reference images in evaluation, the NR-IQA takes an increasingly important role in real-world applications. The traditional NR-IQA methods can be mainly classified into two categories: NSS-based methods \cite{tang2011learning, mittal2012no,mittal2012making,wu2015blind}, and free-energy based methods \cite{friston2006free, friston2010free, gu2013no, gu2014using, zhai2011psychovisual}. For the NSS construction, Tang \textit{et al.} explored the NSS descriptor based on the magnitudes and phases of the pyramid wavelet coefficients \cite{tang2011learning}. A spatial NSS model was constructed by exploring the distribution of locally normalized luminance \cite{mittal2012making}. Xue \textit{et al.} exploited the joint statistics of gradient magnitude map and Laplacian of Gaussian response for NR-IQA \cite{xue2014blind}. In \cite{ye2012unsupervised,ye2012no}, the visual codebooks were constructed from local image patches, aiming for the NSS encoding.  Inspired by the free-energy-based brain theory \cite{friston2006free, friston2010free, gu2013no, gu2014using, zhai2011psychovisual} that HVS always attempts to reduce the uncertainty and explains the perceived scene by an internal generative model, Zhai \textit{et al.} proposed a psychovisual quality model in \cite{zhai2011psychovisual}. Analogously,  Gu \textit{et al.} incorporated the free-energy inspired features and the ``naturalness'' related features for quality regression \cite{gu2013no}. In  \cite{zhang2015som}, the salience map was introduced to guide the quality aggregation due to its high relevance with the semantic obviousness.

Recently, deep learning technologies have achieved great success for NR-IQA. In \cite{kang2014convolutional}, the patch-based NR-IQA  model was learned by several CNN layers. This work was extended by DeepBIQ \cite{bianco2018use}, where a pre-trained network is fine-tuned for the generic image description. In \cite{kang2015simultaneous}, the multi-tasks including the quality prediction and distortion type identification were explored, aiming for the distortion discrimination capability enhancement. However, the deep-learning based methods are usually suffered from the over-fitting problem, due to the limited size of training data. In  \cite{liu2017rankiqa, niu2019siamese,ying2020quality,chen2021no}, the rank information of image pairs was explored which was able to enrich the training samples and mitigate the over-fitting problem to some extent. The generalization performance could also be enhanced by the knowledge transferred from the distortion type identification networks learned from synthesis image databases \cite{zhang2018blind,ma2017end,chen2021no}. In particular, a deep bilinear model was proposed by Zhang \textit{et al.} \cite{zhang2018blind}, combining both the pre-trained image distortion classification network and image category classification network for quality regression. Fang \textit{et al.} \cite{fang2020perceptual} proposed an IQA model for smartphone photography, in which the interactions between image attributes and high-level semantics were explored in a multi-task learning manner. To enhance the model generalization capability on cross-tasks, the incremental learning frameworks were proposed in \cite{ma2023forgetting, ma2021remember}. Chen \textit{et al.} proposed a self-supervised pre-training strategy for quality-aware feature extraction~\cite{chen2022spiq}. Yue \textit{et al.} exploited both labeled data and unlabeled data for a semi-supervised IQA model learning \cite{yue2022semi}.
Inspired by the free-energy theory, the pseudo-reference information was restored in image-level  \cite{lin2018hallucinated} and feature-level \cite{chen2022no}, leading to more discriminative feature extraction.

\subsection{No-reference SCI Quality Assessment Models}
Compared with the NIs, SCIs present distinct statistical differences, leading to a dramatic performance drop when the natural IQA models are tested on SCIs. This phenomenon brings a high demand for IQA models designed specifically for SCIs. In the literature, the NR SCIQA models can be roughly classified into three categories: distortion-aware feature extraction based methods, multi-task learning based methods, and transfer-learning based methods. For the distortion-aware feature extraction, Fang \textit{et al.}  combined the local-global luminance features and texture features with the assumption that HVS was more sensitive to luminance or texture degradation \cite{fang2017no}.  In \cite{gu2017no}, the features that represent picture complexity, screen content statistics, global brightness, and sharpness of details were extracted and regressed for SCI quality prediction. 
The CNNs were utilized for more powerful feature extraction in  \cite{chen2018naturalization,yue2019blind}. For example, Yue \textit{et al.} decomposed the input SCI into predicted and unpredicted portions and learned the quality prediction in an end-to-end manner~\cite{yue2019blind}. For the multi-task learning based methods, Jiang \textit{et al.} learned a  noise classification task auxiliarily for the SCI quality prediction in \cite{jiang2020no}. In \cite{yang2021staged}, the identification tasks of both distortion types and distortion levels were incorporated into the quality prediction model. For the transfer-learning based methods, Chen \textit{et al.} firstly explored the quality assessment transferred from NIs to SCIs with unsupervised domain adaptation~\cite{chen2021no}. The unified IQA models that are able to assess the quality of both NIs and SCIs also attracted much attention \cite{min2018saliency, min2017unified}. For example,  Min \textit{et al.} proposed a unified NR-IQA model with a content-adaptive weighting module \cite{min2017unified}. Though those methods have achieved promising performance on SCIQA, the generalization capability of those models is still not well studied. In this paper,  we make the first attempt to learn the statistics of SCIs and by which, the generalization capability of the SCIQA can be improved significantly.}

\section{The Proposed Scheme}
\begin{figure}[t]
\begin{minipage}[b]{1.0\linewidth}
 \centerline{\includegraphics[width=1\linewidth]{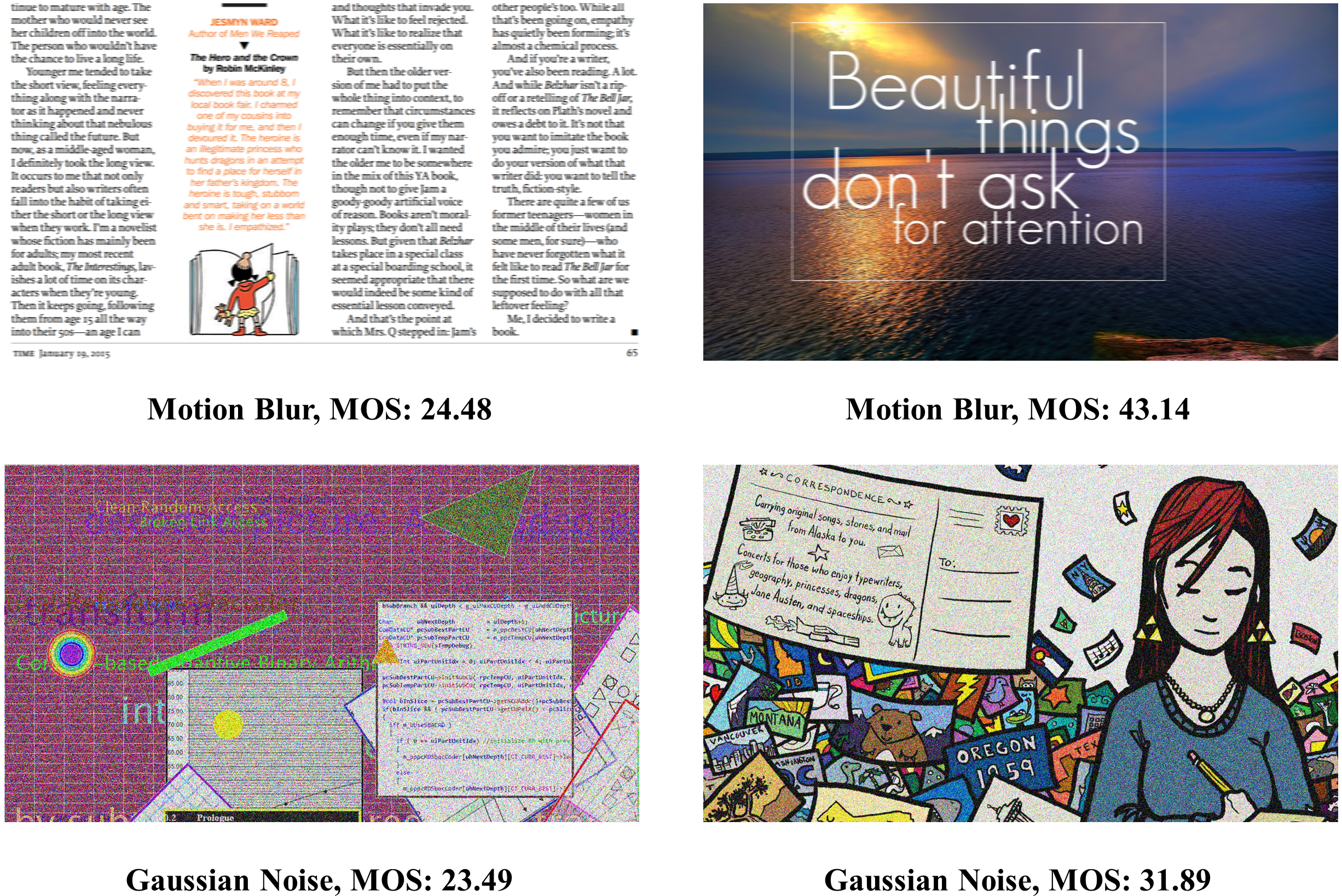}}
\end{minipage}
\caption{\bl{Distorted SCIs sampled from the SCID dataset. First row: Images distorted by the motion blur. Second row: Images distorted by the Gaussian noise. The images in the same row are degraded with the same distortion type and level while possessing different quality scores due to semantic variance.
}}
\label{fig:cmr}
\end{figure}

The goal of the proposed method is to determine the quality of the SCIs without reference. The philosophy behind it is comparing the statistics of the input image against the assumed distribution which is regarded to reflect the ``naturalness'' of pristine SCIs. As such, 
the perceptual quality of SCIs can be estimated by measuring the destruction of the learned distribution. With those mild assumptions, we propose an NR-SCIQA model by constructing the expected distribution in the deep feature space, thus the quality of test images can be regressed from the distribution deviation. As shown in Fig.~\ref{fig:frame}, our proposed framework consists of image patch sampling, quality-aware feature extractor, and quality regression. More specifically, the image patch sampling aims to obtain fixed-size image patches from the original SCI, thus the model can be trained and tested when SCIs are of different resolutions. Subsequently, we extract the quality-aware features of each patch based on a multi-stage feature extractor, from which the distortion at different scales can be effectively captured.  Herein, the quality-aware feature is expected to contain both the semantic information and distortion information of the input SCI. This is reasonable and its evidence can be exemplified in Fig.~\ref{fig:cmr}. From the figure, we can observe that even though the image pairs share the same distortion level, their quality scores can be extremely different, revealing that the image quality is governed by both distortion and content. This phenomenon is also consistent with previous natural IQA works such as \cite{li2017exploiting, zhang2018blind}. Along this vein, we disentangle the quality-aware feature into a semantic-specific feature and a distortion-aware feature. For the distortion-aware feature, we impose a unified distribution on it and treat the predefined distribution as the feature  statistics  of SCIs. Regarding the semantic-specific feature, a triplet constraint is adopted based on the underlying principle that the semantic information of the distorted image is similar to the reference image while possessing a large difference from the auxiliary image.  Finally, the quality of the distorted  SCI can be regressed by synthesizing both the distortion information and semantic information.  
\bl{
Before diving into the details of the proposed model, we first provide the definitions of the symbols in Fig.~\ref{fig:frame}  as follows,
\begin{itemize}
\item $\boldsymbol{I^d, I^r, I^a}$: The input triplet images. In particular, $\boldsymbol{I^d}$ is an image sampled from the distorted images with its corresponding reference image denoted as $\boldsymbol{I^r}$. The $\boldsymbol{I^a}$ is an image sampled from all the reference images while its content is different from the distorted image.
\item $\boldsymbol{F^{qua}}$: The quality-aware feature extracted by the multi-scale feature generator. 

\item $\boldsymbol{F^{rs}$, $F^{ds}$, $F^{as}}$: The semantic-aware features disentangled from the $\boldsymbol{F^{qua}}$. In particular, $\boldsymbol{F^{rs}$, $F^{ds}$, $F^{as}}$ are the semantic-aware feature of $\boldsymbol{I^r, I^d, I^a}$, respectively. Herein,  the semantic-aware feature means the feature that is determined by the image content while is not sensitive to the image distortion, \textit{i.e.}, images with different contents would lead the semantic-specific features to be dissimilar to each other.

\item $\boldsymbol{F^{rd}$, $F^{dd}$, $F^{ad}}$: The distortion-aware  features disentangled from the $\boldsymbol{F^{qua}}$. In particular, $\boldsymbol{F^{rd}$, $F^{dd}$, $F^{ad}}$ are the distortion-aware feature of $\boldsymbol{I^r, I^d, I^a}$, respectively. Herein, the distortion-aware feature means the feature that is sensitive to the image distortion (including the distortion type and distortion level) while not sensitive to the image content.

\item $\boldsymbol{F^{gaus}}$: A feature that is sampled from the pre-defined multivariate normal distribution. It has the same dimension with $\boldsymbol{F^{rd}$, $F^{dd}}$, and $\boldsymbol{F^{ad}}$.

\item $\boldsymbol{\Phi ^{F^{dd}}}$: The KL divergence between the distribution formed by the $\boldsymbol{F^{dd}}$ of different patches in $\boldsymbol{I^d}$ and the multivariate normal distribution.

\item $\boldsymbol{F^{att}}$: A feature generated by $\boldsymbol{F^{ds}}$ which reflects the attention value of each dimension of $\boldsymbol{\Phi ^{F^{dd}}}$.

\item ${Q^{p}}$: The predicted quality score of  $\boldsymbol{I^d}$.

\end{itemize}
}
\subsection{Triplet Samples for Training}
As shown in  Fig.~\ref{fig:frame}, we partition the training samples into triplets and each triplet includes a distorted image ($\boldsymbol{I^{d}}$), its corresponding reference image  ($\boldsymbol{I^{r}}$), and an auxiliary image  ($\boldsymbol{I^{a}}$). In particular, the auxiliary image $\boldsymbol{I^{a}}$ is randomly sampled from the pristine images while containing a different scene with the $\boldsymbol{I^{r}}$. The triplet samples herein aim to disentangle the quality-aware feature into the semantic-specific feature and quality-aware feature which will be elaborated in Sec.~III.B.  In real-world applications, the spatial resolutions of SCIs can be variant, leading to difficulties in training CNN models at the image-level. As a consequence, we crop all the training images into several patches with identical sizes. It should be noted that the triplet images are only utilized in the training phase and only a single test image is needed during testing.

\begin{figure*}[h]
\begin{minipage}[b]{1.0\linewidth}
 \centerline{\includegraphics[width=1.0\linewidth]{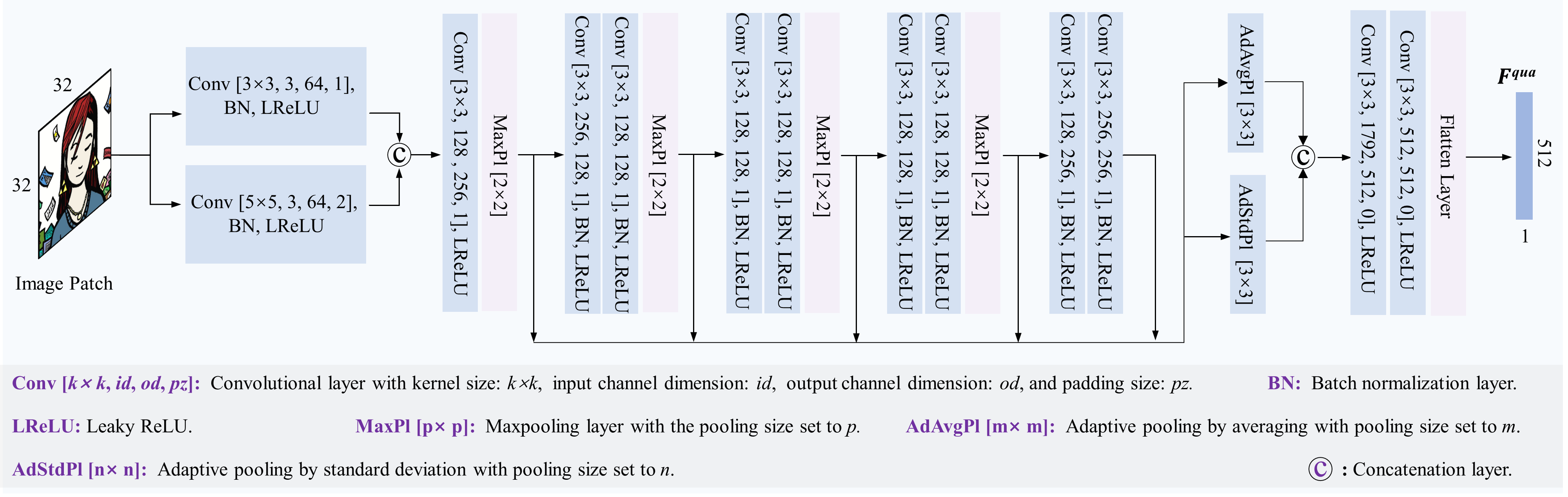}}
\end{minipage}
\caption{{\bl{Structure details of the multi-scale feature generator.} }}
\label{fig:fetgn1}
\end{figure*}

\subsection{Quality-aware Feature Extractor}
The quality-aware feature extractor consists of two stages including the multi-scale feature generator as well as the feature disentanglement. 

\bl{\textbf{1) Multi-scale Feature Generator.} The multi-scale feature generator is constructed based upon the principle that quality-aware features can be well established by the statistics of deep representations at different scales \cite{li2018has, zhang2018unreasonable,bosse2017deep,ding2020image}. We provide the structural details of our multi-scale feature generator in Fig.~4. As shown in the figure, the input of the feature generator is the image patch cropped from the whole image and the feature generator contains five stages at different scales. Each stage of the generator consists of several convolutional layers and one max-pooling layer. The processing of stage $t$ can be formulated as follows,
\begin{equation}
\boldsymbol{F}^{\boldsymbol{ms}}_t=f_t(\boldsymbol{X}),
\end{equation}
where $\boldsymbol{X}$ represents the input patch. $\boldsymbol{F}^{\boldsymbol{ms}}_t$ and $f_t$ mean the output feature map and  processing function at stage $t$ and $t \in \{1,2,3,4,5\}$.  For each $\boldsymbol{F}^{\boldsymbol{ms}}_t$, we adopt the adaptive mean pooling and adaptive standard deviation (std) pooling \cite{he2015spatial} to reduce their spatial dimensions into a fixed scale ($3\times 3$). Herein, the two spatial pooling strategies (mean and std)  have been the widely-used aggregation strategies for IQA. In particular, the mean pooling is able to deliver the global information of learned features \cite{kang2014convolutional,bosse2017deep,zhang2018blind} and the feature variation can be captured by the std pooling \cite{kim2016fully,li2017exploiting,ding2020image,chen2021learning,zhu2022learning}, as such, the two types of pooling strategies are jointly considered in our feature generator. Subsequently, we concatenate  those pooled features along with the channel dimension and further process them by two convolutional layers, leading to the final quality-aware feature $\boldsymbol{F}^{\boldsymbol{qua}}_t$,
\begin{align}
\boldsymbol{F}^{\boldsymbol{qua}}_t  = & f_{mp}(\boldsymbol{F}^{\boldsymbol{ms}}_t) \oplus f_{sp}(\boldsymbol{F}^{\boldsymbol{ms}}_t), t \in \{1,2,3,4,5\}, \\
\boldsymbol{F^{qua}} = & f_c(\boldsymbol{F}^{\boldsymbol{ms}}_1 \oplus \boldsymbol{F}^{\boldsymbol{ms}}_2  \oplus \boldsymbol{F}^{\boldsymbol{ms}}_3 \oplus \boldsymbol{F}^{\boldsymbol{ms}}_4  \oplus \boldsymbol{F}^{\boldsymbol{ms}}_5),
\end{align}
where $f_{mp}$ and $f_{sp}$ represent the adaptive mean pooling and std pooling operations, respectively. $\oplus$ means the concatenation operation. $f_{c}$ is the function of the last two convolutional layers.}

\textbf{ 2) Quality-aware Feature Disentanglement.} As we discussed before, human beings rate the image quality governed by both distortion level as well as the scene semantics. To fully exploit the quality feature of $\boldsymbol{F^{qua}}$, we disentangle it into a distortion-aware feature and a semantic-specific feature. The disentanglement layer consists of two convolutional layers with the ReLU activation function following each.   To ensure the disentanglement of the quality-aware feature $\boldsymbol{{F^{qua}}}$, we first adopt the triplet loss as the objective function with the principle that the semantic features $\boldsymbol{F^{rs}}$ and $\boldsymbol{F^{ds}}$ should be similar to each other while both of them are different from the $\boldsymbol{F^{as}}$ due to the fact that the content  shared by  $\boldsymbol{I^r}$ and $\boldsymbol{I^d}$ are 
different from $\boldsymbol{I^a}$,
\begin{equation}\label{trip}
\begin{aligned}
{\mathcal{L}}^{trip}= \frac{1}{B} \sum_{i=1}^{B}\left[\left\|\boldsymbol{F}^{\boldsymbol{rs}}_i-\boldsymbol{F}^{\boldsymbol{ds}}_i\right\|_{2}^{2}-\left\|\boldsymbol{F}^{\boldsymbol{rs}}_i-\boldsymbol{F}^{\boldsymbol{as}}_i\right\|_{2}^{2}+\alpha\right]_{+},
\end{aligned}
\end{equation}
where $i$ is the input image index in a batch $B$. $\boldsymbol{F}^{\boldsymbol{rs}}_i$, $\boldsymbol{F}^{\boldsymbol{ds}}_i$, and $\boldsymbol{F}^{\boldsymbol{as}}_i$ are the features $\boldsymbol{F}^{\boldsymbol{rs}}$, $\boldsymbol{F}^{\boldsymbol{ds}}$, and $\boldsymbol{F}^{\boldsymbol{as}}$ of $i$-th image. The $\alpha$ is a margin that is enforced between positive and negative pairs.

For the distortion-aware feature, we impose the Gaussian distribution regularization on the feature space. In particular, each dimension of the distortion-aware feature of the pristine SCIs is expected to obey the standard distribution ($mean = 0, std=1$) and is independent of each other. In addition, we force the features of the distorted  SCIs to share the same distribution except that the means and variances are determined by the distortion. To account for this, we first perform the distribution normalization to the distortion-aware feature $\boldsymbol{F^{dd}}$ as follows,

\begin{equation}\label{}
\boldsymbol{F^{dn}} = \frac{ \boldsymbol{F^{dd}}- \boldsymbol{\mu}}{\boldsymbol{\sigma}+\epsilon},  
\end{equation}
where $ \boldsymbol{F^{dn}}$ denotes the normalized results and $\epsilon = 1e-9$ to avoid zero division error. The $ \boldsymbol{\mu}$ and  $ \boldsymbol{\sigma}$ are the mean and std vectors of $\boldsymbol{F^{dd}}$. In particular, supposing $N$ patches are sampled from a distorted image and we denote the $\mu_{j}$ and $\sigma_{j}$  as the mean and std values of the $j$-th dimension of $\boldsymbol{F^{dd}}$. The $\mu_{j}$ and $\sigma_{j}$  can be formulated as follows,
\begin{equation}\label{mu2}
\mu_{j} = \frac{1}{N} \sum_{n=1}^{N}\left(\boldsymbol{F}^{\boldsymbol{dd}}_{n,j} \right),
\end{equation}
and
\begin{equation}\label{mstd2}
\sigma_{j}= \sqrt{\frac{1}{N-1} \sum_{n=1}^{N}(\boldsymbol{F}^{\boldsymbol{dd}}_{n,j}-\mu_{j} )^{2}},\\\\
\end{equation}
where the $\boldsymbol{F}^{\boldsymbol{dd}}_{n,j}$ indicates the $j-$th dimension value of $ \boldsymbol{{F}^{dd}}$ of the $n-$th image patch. Through such normalization, the distribution divergence between different distorted SCIs can be mitigated, and the unified distribution regularization is performed by an MMD loss given by,   
\begin{equation}\label{}
\begin{aligned}
{\mathcal{L}}^{m m d}= \frac{1}{B}  \left\|\sum_{i=1}^{B} \phi\left( \boldsymbol{F}^{\boldsymbol{dn}}_{i}\right)- \sum_{i=1}^{B} \phi\left( \boldsymbol{F}^{\boldsymbol{gaus}}_{i}\right)\right\|_{\mathcal{H}}^{2},
\end{aligned}
\end{equation}
where the $i$ indicates the $i$-th image in a batch $B$  and $ \boldsymbol{{F}^{gaus}}$ is the feature sampled from an independent multivariate normal distribution that shares the same dimensions with $ \boldsymbol{{F}^{dn}}$. The $\phi$ is a function that maps the features into the Reproducing Kernel Hilbert Space (RKHS) \cite{gretton2012kernel}.  We apply the  Gaussian kernel \cite{gretton2012kernel} to compute $\mathcal{L}^{m m d}$, such that the distribution discrepancy between $ \boldsymbol{{F}^{dn}}$ and $ \boldsymbol{{F}^{gaus}}$ is expected to be minimized. Moreover, for the distortion features $ \boldsymbol{{F}^{dd}}$, a distortion type classification loss is further implemented, effectively enhancing the distortion discrimination capability,
\begin{equation}
{\mathcal{L}}^{cls}=-\frac{1}{B}\sum_{i=1}^{B} \sum_{k=1}^{K} y^{k}_{i} \log \left(p^{k}_{i}\right),
\end{equation}
where $i$ indicates the $i$-th input image  in a batch. $p^{k}_{i}$ and $y^{k}_{i}$ indicate the prediction result and ground-truth label of the  $k$-th distortion category, respectively. The $K$ is the total number of distortion categories.

\begin{table*}
  \centering
  \small
  \caption{\bl{Descriptions of the SIQAD and SCID datasets. ``Ref." means reference and ``Dist." means distortion.}}
    \begin{tabular}{c|c|c|m{8.25em}|c|c|m{16.25em}}
    \toprule
    Dataset & \multicolumn{1}{m{4.0em}<{\centering}|}{\# of Ref. Images} & \multicolumn{1}{m{3.5em}<{\centering}|}{\# of Images} & \multicolumn{1}{m{8.25em}<{\centering}|}{Resolution} & \multicolumn{1}{m{5em}<{\centering}|}{\# of Dist. Type } & \multicolumn{1}{m{5em}<{\centering}|}{\# of Dist. Level} & \multicolumn{1}{m{16.25em}<{\centering}}{Dist. Types} \\
    \midrule
    SIQAD & 20    & 980   & {Height Range: 612 to 810  \newline Width Range: \, 624 to 846} & 7     & 7     & Gaussian noise; Gaussian blur; Motion blur; Contrast change; JPEG; JPEG2000; Layer segmentation based coding \\
    \midrule
    SCID  & 40    & 1800  & 1280×720 & 9     & 5     & Gaussian noise; Gaussian blur; Motion blur; Contrast change; JPEG; JPEG2000; Color saturation change; HEVC screen content compression; Color quantization with dithering \\
    \bottomrule
    \end{tabular}%
  \label{tab:dataset}%
\end{table*}%

\subsection{Quality Regression}
The quality of the distorted image is regressed by integrating both the semantic information and the distortion information. In particular, the image distortion is estimated by measuring the distribution deviation between the distribution generated by $N$ sampled image patches and the pre-defined multivariate normal distribution. As both the generated distribution and the normal distribution are Gaussian distributions, their deviation can be efficiently measured by the KL divergence as follows,
\begin{equation}
{ \boldsymbol{\Phi^{F^{dd}}}} = -\frac{1}{2} \left(\boldsymbol{1}+\log  \boldsymbol{\sigma}^{2}-\boldsymbol{\sigma}^{2}- \boldsymbol{\mu}^{2}\right),
\label{eqn:kl}
\end{equation}
where $ \boldsymbol{\mu}$ and $ \boldsymbol{\sigma}$ are mean and std vectors of $ \boldsymbol{{F^{dd}}}$ acquired by Eqn.~\eqref{mu2} and Eqn.~\eqref{mstd2}, respectively. $\boldsymbol{1}$ is a vector with the same dimension as $ \boldsymbol{\sigma}$ and the value of each dimension is 1. We denote the $\boldsymbol{{\Phi}^{F^{dd}}}$ as the distortion feature. To explore the relationship between the semantic feature and distortion feature for final quality regression, we use the semantic feature to generate the channel-wise attention for the distortion feature under the philosophy that the different dimensions of the distortion contribute to final quality with different weights and the weights are determined by the image semantic information. Such processing can be described as follows,
\begin{equation}\label{qp}
Q^{p}= r\left(\varphi ( \boldsymbol{{F}^{ds}}\right) \otimes { \boldsymbol{\Phi^{F^{dd}}}}),
\end{equation}
where $Q^{p}$ is the predicted quality score and the $\varphi(\cdot)$ means the attention generator consists of two fully connected layers with the ReLU and sigmoid as the activation functions, respectively. The $\otimes$ means the point-wise multiplication. The  $r(\cdot)$ represents the regression layer, consisting of a fully connected layer and a ReLU layer. Finally, the regressed quality can be supervised by the mean absolute error (MAE) loss as follows,
\begin{equation}
{\mathcal{L}}^{mae}=\frac{1}{B}\sum_{i=1}^{B} \left\|Q_i^{p}-Q_i^{g}\right\|_1,
\end{equation}
where $Q_i^{g}$ is the ground-truth quality score of the $i-$th input image in a batch. In summary, the final objective function of our model is as follows,
\begin{equation}
\begin{aligned}\label{all}
{\mathcal{L}}^{all}= &{\mathcal{L}}^{mae} + \lambda_1 {\mathcal{L}}^{trip}+ \lambda_2 {\mathcal{L}}^{mmd} + \lambda_3 {\mathcal{L}}^{cls} \\
& + \left(\left\| \boldsymbol{{\Phi}^{F^{rd}}}\right\|_2^2 +\left\| \boldsymbol{{\Phi}^{F^{ad}}}\right\|_2^2+\left\| \boldsymbol{{\Phi}^{F^{rd}}} -\boldsymbol{{\Phi}^{F^{ad}}}\right\|_2^2\right),
\end{aligned}
\end{equation}
where the $  \boldsymbol{{\Phi}^{F^{rd}}}$ and ${ \boldsymbol{\Phi^{F^{ad}}}}$ represent the  distribution divergence of $\boldsymbol{F^{rd}}$ and  $\boldsymbol{F^{ad}}$  measured by Eqn.~\eqref{eqn:kl}. Herein, we minimize their values during training, aiming to persuade them towards a normal distribution. The $ \lambda_1,  \lambda_2$, and $\lambda_3$  are three trade-off parameters.

\section{Experimental Results}
In this section, we first present our experimental setup, including the implementation details of the proposed model and benchmarking datasets. Then we compare the proposed method with the state-of-the-art NR SCI-IQA models in both cross-dataset settings and intra-dataset settings. Next, the ablation study is performed for the functional verification of each module proposed in our method. Finally, we explain the learned distortion-aware and semantic-aware features with significant feature visualizations.

\begin{table*}[h]
  \centering
  \small
  \caption{\bl{Quality prediction results under cross-dataset settings. The top two results are highlighted in boldface.}}
  \resizebox{7.5in}{!}{
  \begin{threeparttable}[b]
    \begin{tabular}{c|ccccccccc}
    \toprule
    \multicolumn{9}{c}{\textbf{Training on SIQAD and Testing on SCID (SIQAD $\rightarrow$ SCID)}} \\
    \midrule
    Method & NIQE  & IL-NIQE & BRISQUE & DIIVINE & CORNIA & HOSA  & BQMS  & SIQE  & ASIQE \\
    \midrule
    SRCC  & 0.2582  & 0.1015  & 0.4526  & 0.4621  & 0.6904  & 0.6024  & 0.6138  & 0.5822  & 0.6053 \\
    PLCC  & 0.3002  & 0.2601  & 0.5050  & 0.4898  & 0.7097  & 0.6147  & 0.6479  & 0.4788  & 0.6630\\
    RMSE  & 13.5089  & 13.6750  & 12.2235  & 12.3468  & 9.9777  & 11.1709  & 10.7787  & 10.9275 & 10.9068 \\
    \midrule
    Method & NRLT  & DIQaM-NR  & WaDIQaM-NR & PQSC  & Bai~\textit{et al.}$^*$  & MtDI & GraphIQA &VCRNet & DFSS-IQA (Ours) \\
    \midrule
    SRCC   & 0.4851  & 0.7151  & 0.5679  & 0.4744  & 0.7294  & 0.6435 &0.7281 & \textbf{0.7605} & \textbf{0.7669 } \\
    PLCC    & 0.5367  & 0.7325 & 0.5775  & 0.4908  & 0.7349  & 0.6419 &0.7348 & \textbf{0.7586} & \textbf{0.7815 } \\
    RMSE    & 11.9496  & 9.7398  & 10.5408   & 12.3392 & \textbf{8.0391}  & -  &9.6058  &8.8751 & \textbf{8.8366}\\
    \midrule
    \multicolumn{9}{c}{\textbf{Training on SCID  and Testing on SIQAD (SCID $\rightarrow$ SIQAD)}} \\
    \midrule
    Method & NIQE  & IL-NIQE & BRISQUE & DIIVINE & CORNIA & HOSA  & BQMS  & SIQE & ASIQE\\
    \midrule
    SRCC  & 0.3594  & 0.3197  & 0.7042  & 0.7065  & \textbf{0.7925 } & 0.7602  & -     & - & - \\
    PLCC  & 0.3805  & 0.3857  & 0.7647  & 0.7762  & 0.8062  & 0.7860  & -     & -  & - \\
    RMSE  & 13.2376  & 13.2065  & 9.2233  & 9.0252  & 8.4682  & 8.8499  & -     & - & - \\
    \midrule
    Method  & NRLT  & DIQaM-NR  & WaDIQaM-NR & PQSC  & Bai~\textit{et al.}  & MtDI & GraphIQA &VCRNet & DFSS-IQA (Ours) \\
    \midrule
    SRCC      & 0.6332  & 0.7507  & 0.7567  & 0.5594  & -  & 0.7611    &0.7811  &0.7815       & \textbf{0.7969 } \\
    PLCC     & 0.6827  & 0.7938  &  0.7988  & 0.6172  & -  & 0.7437   & \textbf{0.8218} &0.8213 & \textbf{0.8344 } \\
    RMSE      & 10.4596  & 8.5806  & 8.5351  & 11.2625  & -  & -   &\textbf{7.9005} & 7.9513  & \textbf{7.8202 } \\
    \bottomrule
    \end{tabular}%
    \begin{tablenotes}
     \item $^*$All the distortion types except for the LSC in the SIQAD dataset are used for training.
   \end{tablenotes}
    \end{threeparttable}
    }
  \label{tab:crossres}%
\end{table*}%

\begin{table*}[h]
  \centering
  \caption{\bbl{Performance on individual distortion types under the cross-dataset setting. The top two results are highlighted in boldface.}}
  \resizebox{7.0in}{!}{
    \begin{tabular}{c|c|p{4.04em}|cccccc|ccc|r}
    \toprule
    \multicolumn{13}{c}{\textbf{Traininig on SIQAD and Testing on SCID}} \\
    \midrule
    \multicolumn{3}{c|}{\multirow{2}[4]{*}{Distortion Type}} & \multicolumn{6}{c|}{Shared }                  & \multicolumn{3}{c|}{Unseen} & \multicolumn{1}{c}{\multirow{2}[4]{*}{Overall}} \\
\cmidrule{4-12}    \multicolumn{3}{c|}{} & {GN} & {GB} & {MB} & {CC} & {JPEG } & {J2K} & {CSC} & {HEVC-SCC} &{CQD} &  \\
    \midrule
    \multirow{10}[10]{*}{Traditional} & \multicolumn{1}{c|}{\multirow{2}[2]{*}{CORNIA}} & SRCC  & 0.7165  & \textbf{0.8019 } & 0.6903  & \textbf{0.7359 } & 0.6081  & 0.7324  & \textbf{0.4821 } & 0.1391  & 0.3372  & 0.6904 \\
          &       & PLCC  & 0.7104  & \textbf{0.8096 } & 0.7049  & \textbf{0.7946 } & 0.6439  & 0.7423  & \textbf{0.6876 } & 0.2434  & 0.3825  & 0.7097 \\
\cmidrule{2-13}          & \multicolumn{1}{c|}{\multirow{2}[2]{*}{HOSA}} & SRCC  & 0.6020  & 0.6646  & 0.7383  & 0.6109  & 0.4244  & 0.3944  & \textbf{0.4278 } & 0.1383  & 0.4844  & 0.6024 \\
          &       & PLCC  & 0.6259  & 0.7130  & 0.7606  & 0.7157  & 0.4463  & 0.4427  & \textbf{0.6665 } & 0.3263  & \textbf{0.5206 } & 0.6147 \\
\cmidrule{2-13}          & \multicolumn{1}{c|}{\multirow{2}[2]{*}{NRLT}} & SRCC  & 0.8604  & 0.4571  & 0.6079  & 0.0721  & 0.5634  & 0.4953  & 0.1370  & 0.1816  & 0.1033  & 0.4851 \\
          &       & PLCC  & 0.8602  & 0.4593  & 0.6241  & 0.1284  & 0.5716  & 0.5198  & 0.1896  & 0.1902  & 0.1279  & 0.5367 \\
\cmidrule{2-13}          & \multicolumn{1}{c|}{\multirow{2}[2]{*}{BQMS}} & SRCC  & \textbf{0.9437 } & 0.7308  & 0.5378  & 0.3548  & 0.6651  & 0.5610  & 0.0064  & 0.4203  & 0.3454  & 0.6138 \\
          &       & PLCC  & \textbf{0.9499 } & 0.7336  & 0.5639  & 0.3896  & 0.6846  & 0.5871  & 0.0106  & 0.5574  & 0.4672  & 0.6479 \\
\cmidrule{2-13}          & \multicolumn{1}{c|}{\multirow{2}[2]{*}{PQSC}} & SRCC  & 0.8125  & 0.5365  & 0.5744  & 0.4699  & 0.4407  & 0.1340  & 0.0664  & 0.2960  & 0.1647  & 0.4744 \\
          &       & PLCC  & 0.8405  & 0.5741  & 0.5865  & 0.5251  & 0.4753  & 0.1731  & 0.1789  & 0.3442  & 0.3201  & 0.4908 \\
    \midrule
    \multirow{10}[10]{*}{Deep Learning} & \multicolumn{1}{c|}{\multirow{2}[2]{*}{WaDIQaM-NR}} & SRCC  & \textbf{0.9279 } & 0.4732  & 0.3995  & 0.5297  & 0.4892  & 0.5075  & 0.1618  & 0.2661  & \textbf{0.5121 } & 0.5679 \\
          &       & PLCC  & \textbf{0.9266 } & 0.4540  & 0.4174  & 0.6281  & 0.5205  & 0.4960  & 0.1739  & 0.3107  & 0.5183  & 0.5775 \\
\cmidrule{2-13}          & \multicolumn{1}{c|}{\multirow{2}[2]{*}{DIQaM-NR}} & SRCC  & 0.9088  & 0.7800  & \textbf{0.7881 } & 0.4773  & 0.7518  & 0.7471  & 0.1683  & 0.4008  & 0.4910  & 0.7151 \\
          &       & PLCC  & 0.9105  & 0.7878  & \textbf{0.7892 } & 0.7225  & 0.7515  & 0.7570  & 0.1832  & 0.4366  & 0.4867  & 0.7325 \\
\cmidrule{2-13}          & \multicolumn{1}{c|}{\multirow{2}[2]{*}{GraphIQA}} & SRCC  & 0.8748  & 0.7710  & 0.6241  & \textbf{0.6240 } & 0.7595  & 0.7884  & 0.2180  & \textbf{0.5024 } & 0.4028  & 0.7281 \\
          &       & PLCC  & 0.8937  & 0.7676  & 0.6084  & 0.6686  & 0.7590  & 0.7929  & 0.2193  & \textbf{0.5165 } & 0.4736  & 0.7348 \\
\cmidrule{2-13}          & \multicolumn{1}{c|}{\multirow{2}[2]{*}{VCRNet}} & SRCC  & 0.8943  & 0.7803  & 0.7592  & 0.5493  & \textbf{0.7775 } & \textbf{0.7943 } & 0.1750  & 0.4856  & 0.4112  & \textbf{0.7605} \\
          &       & PLCC  & 0.8717  & 0.7643  & 0.7649  & 0.5919  & \textbf{0.7847 } & \textbf{0.8024 } & 0.1811  & 0.5086  & 0.4299  & \textbf{0.7586} \\
\cmidrule{2-13}          & \multicolumn{1}{c|}{\multirow{2}[2]{*}{DFSS-IQA (Ours)}} & SRCC  & 0.8934  & \textbf{0.8554 } & \textbf{0.8122 } & 0.5546  & \textbf{0.8707 } & \textbf{0.8495 } & 0.1929  & \textbf{0.5714 } & \textbf{0.4957 } & \textbf{0.7669} \\
          &       & PLCC  & 0.9086  & \textbf{0.8560 } & \textbf{0.8186 } & \textbf{0.7530 } & \textbf{0.8795 } & \textbf{0.8623 } & 0.2301  & \textbf{0.5977 } & \textbf{0.5429 } & \textbf{0.7815} \\
    \bottomrule
    \end{tabular}%
    }
  \label{tab:crosseach}%
\end{table*}%

\begin{table*}[h]
\small 
  \centering
  \caption{\bbl{Performance on individual distortion types under the cross-dataset setting. The top two results are highlighted in boldface.}}
  \resizebox{7.0in}{!}{
    \begin{tabular}{c|c|p{4.04em}|cccccc|c|c}
    \toprule
    \multicolumn{11}{c}{\textbf{Traininig on SCID and Testing on SIQAD}} \\
    \midrule
    \multicolumn{3}{c|}{\multirow{2}[4]{*}{Distortion Type}} & \multicolumn{6}{c|}{Shared }                  & Unseen & \multirow{2}[4]{*}{Overall} \\
\cmidrule{4-10}    \multicolumn{3}{c|}{} & {GN} & {GB} & {MB} & {CC} & {JPEG } & {J2K} & {LSC} &  \\
    \midrule
    \multirow{8}[8]{*}{Traditional} & \multicolumn{1}{c|}{\multirow{2}[2]{*}{CORNIA}} & SRCC  & 0.7675  & 0.7879  & 0.6843  & \textbf{0.6606 } & 0.5490  & 0.6283  & \textbf{0.7023 } & \textbf{0.7925} \\
          &       & PLCC  & 0.7862  & 0.8077  & 0.6805  & \textbf{0.8233 } & 0.5710  & 0.6405  & \textbf{0.7956 } & 0.8062 \\
\cmidrule{2-11}          & \multicolumn{1}{c|}{\multirow{2}[2]{*}{HOSA}} & SRCC  & 0.7873  & 0.8334  & 0.8209  & 0.4382  & 0.7013  & 0.5546  & \textbf{0.7530 } & 0.7602 \\
          &       & PLCC  & 0.8179  & \textbf{0.8940 } & \textbf{0.8131 } & 0.6987  & 0.7246  & 0.5811  & \textbf{0.7756 } & 0.7860 \\
\cmidrule{2-11}          & \multicolumn{1}{c|}{\multirow{2}[2]{*}{NRLT}} & SRCC  & 0.8800  & 0.6596  & 0.6091  & 0.0467  & \textbf{0.7472 } & 0.2663  & 0.4982  & 0.6332 \\
          &       & PLCC  & 0.8903  & 0.6653  & 0.6146  & 0.0143  & \textbf{0.7511 } & 0.2840  & 0.5223  & 0.6827 \\
\cmidrule{2-11}          & \multicolumn{1}{c|}{\multirow{2}[2]{*}{PQSC}} & SRCC  & 0.8540  & 0.7062  & 0.4534  & 0.0431  & 0.4584  & 0.4034  & 0.4024  & 0.5594 \\
          &       & PLCC  & 0.8684  & 0.7538  & 0.4504  & 0.0086  & 0.4378  & 0.4833  & 0.4061  & 0.6172 \\
    \midrule
    \multirow{10}[10]{*}{Deep Learning} & \multicolumn{1}{c|}{\multirow{2}[2]{*}{WaDIQaM-NR}} & SRCC  & 0.8845  & 0.8404  & 0.7876  & 0.5201  & 0.6250  & 0.5472  & 0.4216  & 0.7567 \\
          &       & PLCC  & 0.8833  & 0.8292  & 0.7913  & 0.6811  & 0.6093  & 0.5764  & 0.4455  & 0.7988 \\
\cmidrule{2-11}          & \multicolumn{1}{c|}{\multirow{2}[2]{*}{DIQaM-NR}} & SRCC  & \textbf{0.8936 } & 0.8326  & 0.7698  & 0.3834  & 0.5919  & 0.5403  & 0.3750  & 0.7507 \\
          &       & PLCC  & 0.8676  & 0.8491  & 0.7812  & 0.6664  & 0.6204  & 0.5732  & 0.4147  & 0.7938 \\
\cmidrule{2-11}          & \multicolumn{1}{c|}{\multirow{2}[2]{*}{GraphIQA}} & SRCC  & 0.8840  & \textbf{0.8767 } & 0.8232  & \textbf{0.5574 } & 0.6066  & 0.6072  & 0.4525  & 0.7811 \\
          &       & PLCC  & \textbf{0.8912 } & \textbf{0.8780 } & 0.8095  & \textbf{0.7240 } & 0.6175  & 0.6092  & 0.4138  & \textbf{0.8218} \\
\cmidrule{2-11}          & \multicolumn{1}{c|}{\multirow{2}[2]{*}{VCRNet}} & SRCC  & 0.8773  & \textbf{0.8671 } & \textbf{0.8467 } & 0.5029  & 0.7091  & \textbf{0.6618 } & 0.5473  & 0.7815 \\
          &       & PLCC  & 0.8649  & 0.8564  & 0.8077  & 0.6433  & 0.6842  & \textbf{0.6613 } & 0.5474  & 0.8213 \\
\cmidrule{2-11}          & \multicolumn{1}{c|}{\multirow{2}[2]{*}{DFSS-IQA (Ours)}} & SRCC  & \textbf{0.8953 } & 0.8443  & \textbf{0.8249 } & 0.4921  & \textbf{0.7158 } & \textbf{0.6484 } & 0.6216  & \textbf{0.7969} \\
          &       & PLCC  & \textbf{0.8999 } & 0.8487  & \textbf{0.8368 } & 0.6954  & \textbf{0.7443 } & \textbf{0.6619 } & 0.6480  & \textbf{0.8344} \\
    \bottomrule
    \end{tabular}%
    }
  \label{tab:cross-siqadeach}%
\end{table*}%

\subsubsection{Implementation Details}
We implement  our  model  by  PyTorch~\cite{paszke2019pytorch}. In the training phase, we crop the image into patches without overlapping with the patch size set by $32\times 32$. The number of images in a batch is 96, including 32 distorted images, 32 reference images, and 32 auxiliary images (as shown in Fig.~\ref{fig:frame}). 
We adopt Adam optimizer~\cite{kingma2014adam} for optimization. The learning rate is fixed to 1e-4 with a weight decay set as 1e-4. The  parameters $\lambda_1$, $\lambda_2$ and $\lambda_3$ in Eqn.~\eqref{all} are set by 1.0, 5e-3 and 1.0, respectively. We duplicate the samples 16 times in a batch to augment the data. The maximum epoch is set by 200.  The $\alpha$ in  Eqn.~\eqref{trip} is a positive value that decides the margin between positive pairs and negative pairs. We empirically set its value to 1.0 during training.

It is worth mentioning that all the experimental pre-settings are fixed in both intra-database and cross-database training. For the intra-database evaluation, we randomly split the dataset into a training set, a validation set, and a testing set by reference images to guarantee there is no content overlap among the three subsets. In particular, 60\%, 20\%, and 20\% images are used for training, validation, and testing, respectively.  The experimental results on intra-database are reported based on 10 random splits. To make errors and gradients comparable for different databases, we linearly map the MOS/DMOS ranges of the SIQAD and SCID databases to the DMOS range [0, 100]. Three evaluation metrics are reported for each experimental setting, including Spearman Rank Correlation Coefficient (SRCC), Pearson Linear Correlation Coefﬁcient (PLCC), and Root Mean Square Error (RMSE). The  PLCC  estimates the prediction linearity and consistency, SRCC measures the prediction monotonicity, and RMSE  evaluates the prediction accuracy.  Higher PLCC, SRCC, and lower RMSE indicate better performance of the quality model.

\subsubsection{SCI Datasets}
\bl{More details regarding the SIQAD dataset and SCID dataset can be found in  Table \ref{tab:dataset}}, which are briefly introduced as follows.
\begin{itemize}
\item SIQAD dataset \cite{yang2015perceptual} contains 20 reference SCIs and 980 distorted SCIs. The distorted images are derived from seven distortion types including Gaussian Noise (GN), Gaussian Blur (GB), Motion Blur (MB), Contrast Change (CC), JPEG, JPEG2000, and Layer Segmentation based Coding (LSC). For each distortion type, seven distortion levels are generated. 

\item SCID dataset \cite{ni2017scid} consists of 1800 distorted SCIs generated by 40 reference images. In this dataset,  nine distortion types are involved including  GN, GB, MB, CC, JPEG, JPEG2000, Color Saturation Change (CSC),  High-Efficiency Video Coding Screen Content Compression (HEVC-SCC), and Color Quantization with Dithering (CQD). Each distortion type contains five degradation levels.  All the SCIs in SIQAD are with a resolution of 1280 × 720.
\end{itemize}

\subsection{Quality Prediction on Cross-dataset Settings}
In this subsection, we first evaluate the performance of our method on the SIQAD and SCID datasets with the cross-dataset settings. 
Herein, there are two settings: training on the SIQAD  dataset and testing on the SCID dataset, and training on the SCID  dataset and testing on the SIQAD dataset. These settings are denoted as SIQAD $\rightarrow$ SCID and  SCID $\rightarrow$ SIQAD, respectively.
There exists a large domain gap caused by the unshared content, distortion types, and distortion levels between two different datasets, such that cross-dataset testing appears to be an effective way to evaluate the model generalization capability. We compare the proposed method with both handcrafted feature based methods including NIQE~\cite{mittal2012making},  IL-NIQE~\cite{zhang2015feature},  BRISQUE~\cite{mittal2012no}, DIIVINE~\cite{moorthy2011blind},    CORNIA~\cite{ye2012unsupervised}, HOSA~\cite{xu2016blind}, BQMS~\cite{gu2016learning},  SIQE~\cite{gu2017no},	ASIQE~\cite{gu2017no},	NRLT~\cite{fang2017no}, and deep-learning based methods including  DIQaM-NR~\cite{bosse2017deep}, WaDIQaM-NR~\cite{bosse2017deep}, PQSC~\cite{fang2019perceptual}, RIQA~\cite{jiang2020no}, MtDI~\cite{yang2021staged} GraphIQA \cite{sun2022graphiqa}, and VCRNet \cite{pan2022vcrnet}. 
As shown in Table~\ref{tab:crossres}, most deep-learning based methods achieve inferior performance than the traditional ones (\eg, CORNIA), revealing the so-called over-fitting problem. However, our method achieves the best overall performance in both two settings, demonstrating superior generalization capability that can be learned by the scene statics construction. In Table~\ref{tab:crosseach} and Table~\ref{tab:cross-siqadeach}, we further present the performance comparison on the individual distortion types. In Table Table~\ref{tab:crosseach},  the distortion types including CC, HEVC-SCC, and CQD in the SCID dataset are unseen when trained on the SIQAD dataset, leading to a significant performance drop compared with other distortion types. However, our method still achieves superior performance on those distortion types when compared with other deep-learning based models. A consistent phenomenon can be observed in the setting Table Table~\ref{tab:cross-siqadeach} (SCID $\rightarrow$ SIQAD). Herein, it should be noted that our method can only alleviate but not completely eliminate the challenges posed by unseen data, motivating further exploration of high generalization models based upon the statistics learning.

\begin{table*}[h]
  \centering
  \caption{{Performance on SIQAD dataset under the intra-dataset setting. The top two results are highlighted in boldface.}}
  \resizebox{7in}{!}{
    \begin{tabular}{c|c|ccccccccccccccc|c}
    \toprule
    \multicolumn{1}{c|}{Criteria} & Distortion & \multicolumn{1}{c}{NIQE} & \multicolumn{1}{c}{BRISQUE} & \multicolumn{1}{c}{QAC} & \multicolumn{1}{c}{IL-NIQE} & \multicolumn{1}{c}{BQMS} & \multicolumn{1}{c}{SIQE} & \multicolumn{1}{c}{ASIQE} & \multicolumn{1}{c}{NRLT} & \multicolumn{1}{c}{HRFF} & \multicolumn{1}{c}{CLGF} & Li~\textit{et al.} & \multicolumn{1}{c}{DIQaM-NR} & \multicolumn{1}{c}{WaDIQaM-NR} & \multicolumn{1}{c}{Yang~\textit{et al.} (TIP21)} & \multicolumn{1}{c|}{ Yang~\textit{et al.} (Tcy20)} & \multicolumn{1}{c}{DFSS-IQA (Ours)} \\
    \midrule
    \multicolumn{1}{c|}{\multirow{8}[4]{*}{PLCC}} & GN    & 0.8339  & 0.8346  & 0.8525  & 0.7156  & 0.8377  & 0.8779  & 0.8398  & 0.9131  & 0.9020  & 0.8577  & \multicolumn{1}{c}{0.8957 } & \textbf{0.9270 } & 0.9158  & 0.9139  & \textbf{0.9249 } & 0.9245  \\
          & GB    & 0.5946  & 0.8359  & 0.5587  & 0.5238  & 0.8739  & 0.9138  & 0.9059  & 0.8949  & 0.8900  & 0.9082  & \multicolumn{1}{c}{0.9164 } & 0.9114  & 0.8990  & \textbf{0.9225 } & \textbf{0.9229 } & 0.9164  \\
          & MB    & 0.2878  & 0.7666  & 0.3780  & 0.4657  & 0.6733  & 0.7836  & 0.7724  & \textbf{0.8993 } & 0.8740  & 0.8609  & \multicolumn{1}{c}{0.8679 } & 0.8207  & 0.8869  & 0.8948  & \textbf{0.8980 } & 0.8825  \\
          & CC    & 0.3132  & 0.5489  & 0.0744  & 0.1098  & 0.3146  & 0.6856  & 0.6894  & 0.8131  & 0.8260  & 0.7440  & \multicolumn{1}{c}{0.8129 } & \textbf{0.8609 } & 0.8245  & 0.7772  & 0.7823  & \textbf{0.8563 } \\
          & JPEG  & 0.4591  & 0.7346  & 0.3017  & 0.3296  & 0.6096  & 0.7244  & 0.6756  & 0.7932  & 0.7630  & 0.6598  & \multicolumn{1}{c}{0.8326 } & 0.7858  & \textbf{0.8491 } & 0.8014  & 0.8115  & \textbf{0.8383 } \\
          & J2K   & 0.3774  & 0.7645  & 0.1885  & 0.4184  & 0.6358  & 0.7339  & 0.6381  & 0.6848  & 0.7540  & 0.7463  & \multicolumn{1}{c}{\textbf{0.9279 }} & 0.7683  & 0.8077  & 0.7984  & \textbf{0.8734 } & 0.7946  \\
          & LSC   & 0.4314  & 0.6980  & 0.3367  & 0.1502  & 0.4814  & 0.7332  & 0.6413  & 0.7228  & 0.7700  & 0.5575  & \multicolumn{1}{c}{\textbf{0.7999 }} & 0.7536  & 0.7842  & \textbf{0.7907 } & 0.7460  & 0.7844  \\
\cmidrule{2-18}          & ALL   & 0.3415  & 0.7237  & 0.3751  & 0.3854  & 0.7575  & 0.7906  & 0.7884  & 0.8442  & 0.8520  & 0.8331  & \multicolumn{1}{c}{\textbf{0.8834 }} & 0.8799  & \textbf{0.8890 } & 0.8529  & 0.8738  & 0.8818  \\
    \midrule
    \multicolumn{1}{c|}{\multirow{8}[4]{*}{SRCC}} & GN    & 0.8324  & 0.8665  & 0.8416  & 0.7502  & 0.8347  & 0.8517  & 0.8299  & 0.8966  & 0.8720  & 0.8478  & \multicolumn{1}{c}{0.9075 } & 0.8993  & 0.9070  & 0.9102  & \textbf{0.9180 } & \textbf{0.9105 } \\
          & GB    & 0.6178  & 0.8234  & 0.6238  & 0.5034  & 0.8591  & \textbf{0.9174 } & 0.9021  & 0.8812  & 0.8630  & 0.9152  & \multicolumn{1}{c}{0.9150 } & 0.8810  & 0.8886  & \textbf{0.9223 } & 0.9163  & 0.8987  \\
          & MB    & 0.3921  & 0.6786  & 0.3375  & 0.4253  & 0.6707  & 0.8347  & 0.7765  & \textbf{0.8919 } & 0.8500  & 0.8694  & \multicolumn{1}{c}{0.8674 } & 0.8134  & 0.8588  & 0.8867  & \textbf{0.8931 } & 0.8675  \\
          & CC    & 0.1702  & 0.7256  & 0.0745  & 0.0402  & 0.2450  & 0.6874  & 0.4068  & 0.7072  & 0.6870  & 0.5716  & \multicolumn{1}{c}{\textbf{0.7952 }} & 0.7745  & 0.7274  & 0.7471  & 0.7785  & \textbf{0.7953 } \\
          & JPEG  & 0.4467  & 0.7543  & 0.1451  & 0.2745  & 0.6026  & 0.7438  & 0.6624  & 0.7698  & 0.7180  & 0.6778  & \multicolumn{1}{c}{0.7863 } & 0.7441  & 0.8027  & 0.7768  & \textbf{0.8081 } & \textbf{0.8142 } \\
          & J2K   & 0.3558  & 0.7456  & 0.1937  & 0.3880  & 0.6182  & 0.7241  & 0.6241  & 0.6761  & 0.7440  & 0.7681  & \multicolumn{1}{c}{\textbf{0.8804 }} & 0.7427  & 0.7608  & 0.7783  & \textbf{0.8669 } & 0.7499  \\
          & LSC   & 0.3953  & 0.6323  & 0.1866  & 0.1535  & 0.5215  & 0.7337  & 0.6216  & 0.6978  & 0.7400  & 0.5842  & \multicolumn{1}{c}{\textbf{0.8116 }} & 0.7028  & \textbf{0.7786 } & 0.7585  & 0.7355  & 0.7510  \\
\cmidrule{2-18}          & ALL   & 0.3695  & 0.7708  & 0.3009  & 0.3575  & 0.7251  & 0.7625  & 0.7570  & 0.8202  & 0.8320  & 0.8107  & \multicolumn{1}{c}{0.8645 } & 0.8662  & \textbf{0.8780 } & 0.8336  & 0.8543  & \textbf{0.8820 } \\
    \midrule
    \multicolumn{1}{c|}{\multirow{8}[4]{*}{RMSE}} & GN    & 8.2319  & 7.5643  & 8.1054  & 8.4529  & 8.1451  & 8.1416  & 8.0975  & 6.2678  & 6.3110  & \multicolumn{1}{c}{-} & -     & \textbf{5.5676 } & 5.9814  & 5.9745  & 6.3625  & \textbf{5.5456 } \\
          & GB    & 7.9880  & 8.4593  & 7.8821  & 9.5842  & 7.3769  & 6.4239  & 6.4267  & 6.7385  & 6.9170  & \multicolumn{1}{c}{-} & -     & 6.2084  & 6.7192  & \textbf{5.7319 } & 5.7767  & \textbf{5.7077 } \\
          & MB    & 10.6760  & 9.4532  & 10.1330  & 10.7232  & 9.6127  & 8.0783  & 8.2582  & 6.4600  & 6.4520  & \multicolumn{1}{c}{-} & -     & 6.9046  & \textbf{5.5435 } & 6.7144  & \textbf{5.6846 } & 5.9666  \\
          & CC    & 10.0420  & 8.6786  & 12.3041  & 12.9043  & 11.9399  & 9.1565  & 9.1116  & 7.8744  & 7.8430  & \multicolumn{1}{c}{-} & -     & \textbf{6.2280 } & 7.0092  & 8.0684  & 7.5985  & \textbf{5.8874 } \\
          & JPEG  & 8.7865  & 5.8643  & 11.0892  & 9.4592  & 7.4485  & 6.4778  & 6.9279  & 5.8625  & 5.8720  & \multicolumn{1}{c}{-} & -     & 6.5162  & \textbf{5.2383 } & 6.8006  & \textbf{5.3835 } & 5.7826  \\
          & J2K   & 8.3476  & 6.1375  & 10.9169  & 8.4099  & 8.0220  & 7.6727  & 8.0021  & 6.5017  & 6.5440  & \multicolumn{1}{c}{-} & -     & 6.2439  & 5.6635  & 6.5538  & \textbf{4.8967 } & \textbf{5.4014 } \\
          & LSC   & 7.6972  & 6.2321  & 11.4067  & 7.9421  & 7.4781  & 6.3160  & 6.5465  & 5.4734  & 5.7860  & \multicolumn{1}{c}{-} & -     & 5.6322  & 5.5285  & \textbf{5.4556 } & 5.8593  & \textbf{5.3568 } \\
\cmidrule{2-18}          & ALL   & 13.4670  & 8.2565  & 13.2690  & 13.9320  & 9.3456  & 8.7650  & 8.8064  & 7.4156  & 7.5960  & 7.9172  & -     & 6.7894  & \textbf{6.7552 } & 7.2817  & 6.9335  & \textbf{6.6684 } \\
    \bottomrule
    \end{tabular}%
    }
  \label{tab:siqad1}%
\end{table*}%

\subsection{Quality Prediction on Intra-dataset Settings}
In this sub-section, \bl{we first compare our method with several state-of-the-art NR-IQA methods on the SIQAD dataset. The  competing models include NIQE~\cite{mittal2012making}, BRISQUE~\cite{mittal2012no}, QAC~\cite{xue2013learning}, IL-NIQE~\cite{zhang2015feature}, BQMS~\cite{gu2016learning},  SIQE~\cite{gu2017no},	ASIQE~\cite{gu2017no},	NRLT~\cite{fang2017no},	HRFF~\cite{zheng2019no},	CLGF~\cite{wu2019blind}, Li~\textit{et al.}~\cite{li2019cnn}, DIQaM-NR~\cite{bosse2017deep}, WaDIQaM-NR~\cite{bosse2017deep}, Yang~\textit{et al.} (Tcy20)~\cite{yang2020no}, Yang~\textit{et al.} (TIP21)~\cite{yang2021no}, GraphIQA \cite{sun2022graphiqa}, and VCRNet \cite{pan2022vcrnet}. } The intra-dataset experiment is conducted under ten times random splitting and the median SRCC and PLCC values are reported as final statistics in Table~\ref{tab:siqad1} and Table~\ref{tab:siqad}. In these tables, both the overall performance and the performance of individual distortion types are presented. The top two results are emphasized in boldface. From Table~\ref{tab:siqad}, we can observe that our method achieves the best overall performance on the  SIQAD dataset in terms of both PLCC and RMSE. The SRCC results are also comparable with the best method  WaDIQaM-NR (0.8818 \textit{v.s.} 0.8890), revealing that the high generalization capability our method could achieve without the sacrifice of intra-dataset performance. In addition, our method presents an effective quality assessment for each individual distortion type as the distortion-aware features can be learned by measuring the KL-divergence. 
\begin{figure}[t]
\begin{minipage}[b]{1.0\linewidth}
 \centerline{\includegraphics[width=0.90\linewidth]{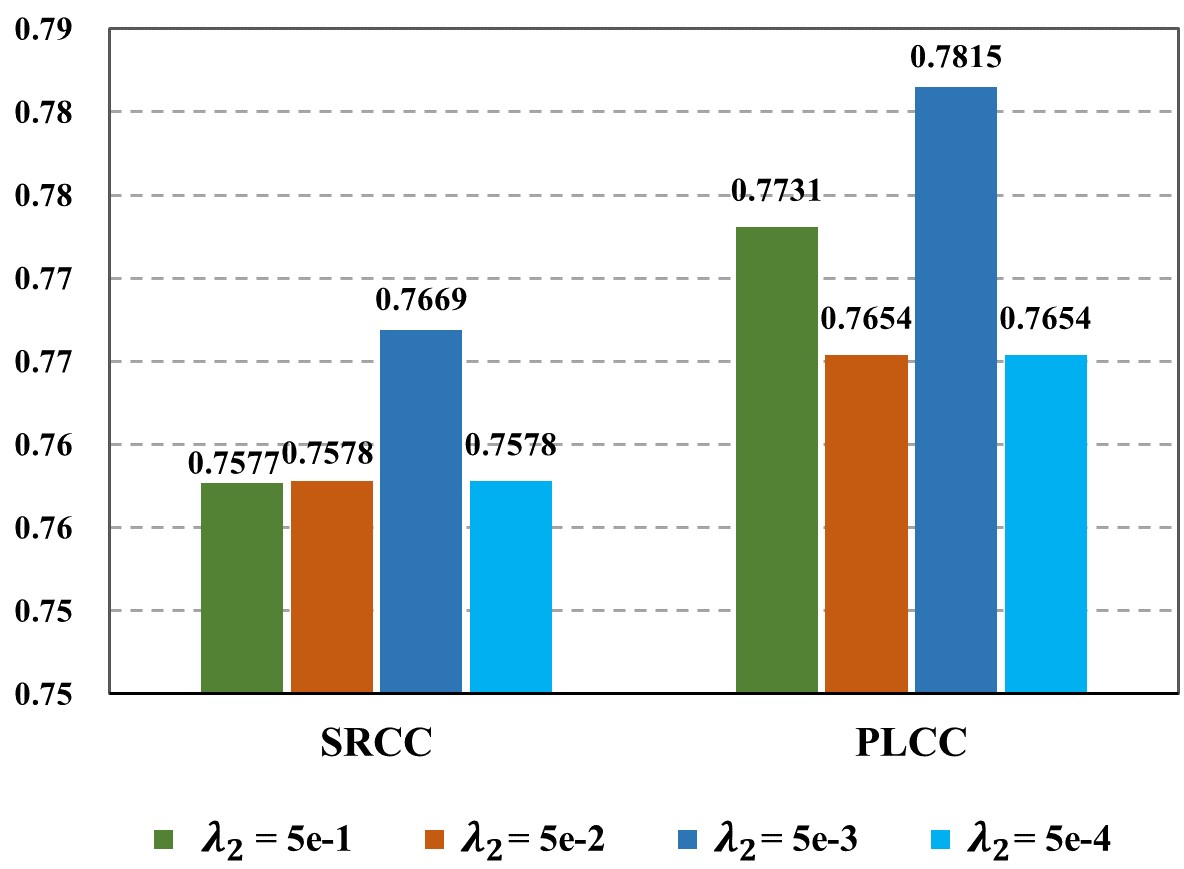}}
\end{minipage}
\caption{{Study of  the hyperparameter $\lambda_2$ under the cross-dataset setting: SIQAD $\rightarrow$ SCID.}}
\label{fig:lamda}
\end{figure}
To further demonstrate the effectiveness of our method, we also conduct the intra-dataset experiment on the SCID dataset. The experimental results are shown in Table~\ref{tab:scid}, from which we can observe our method achieves the best performance in terms of RMSE, while its performance of PLCC and SRCC is slightly inferior to GraphIQA and VCRNet.  It is worth noting that both GraphIQA and VCRNet are retrained on large-scale IQA datasets~\cite{2019kadid,ma2016waterloo} and ImageNet~\cite{deng2009imagenet}. As such, it is not surprising that they demonstrate inferior performance with the cross-dataset setting (see Table II). In contrast, our method, which computes the statistics in the deep feature domain, shows superior performance and strong generalization capabilities in both intra-dataset and cross-dataset settings. 
\begin{figure*}[t]
\begin{minipage}[b]{1.0\linewidth}
 \centerline{\includegraphics[width=0.8\linewidth]{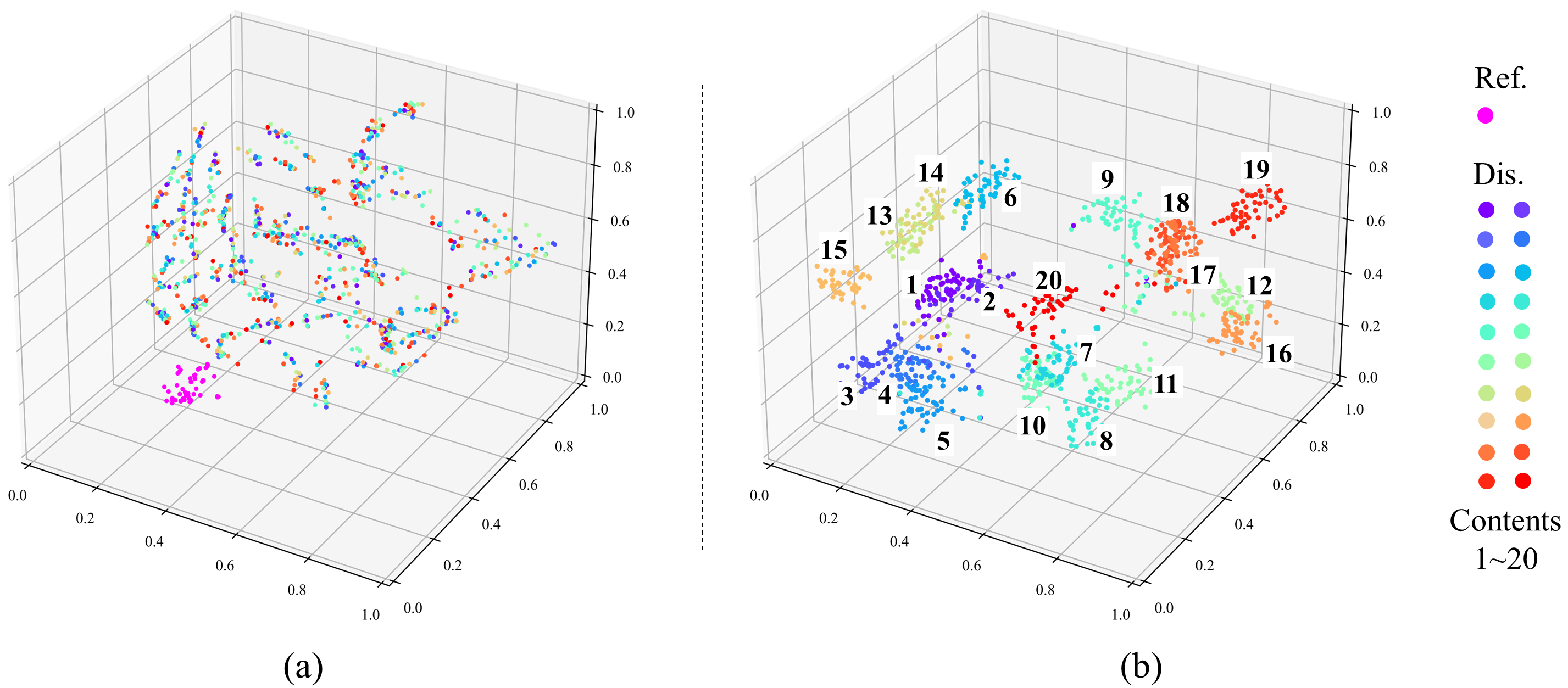}}
\end{minipage}
\caption{\bbl{T-SNE~\cite{maaten2008visualizing}  visualization of the disentangled features. (a)  Distortion-aware features ($\boldsymbol{\Phi ^{F^{rd}}}$ and $\boldsymbol{\Phi ^{F^{dd}}}$). (b) Semantic-aware features ($\boldsymbol{\Phi ^{F^{ds}}}$). The ``Ref.'' and ``Dis.''  mean the features extracted from reference images and distorted images, respectively. The distortion-aware features of the reference images are clustered into one group due to their similar quality and the features of distorted images are distributed in different spaces by their distortion types and levels.  In comparison, the semantic-specific features are clustered into 20 groups as different content types those images contain. }}
\label{fig:tsne}
\end{figure*}

\begin{figure*}[!htb]
  \centerline{\includegraphics[width=0.9\linewidth]{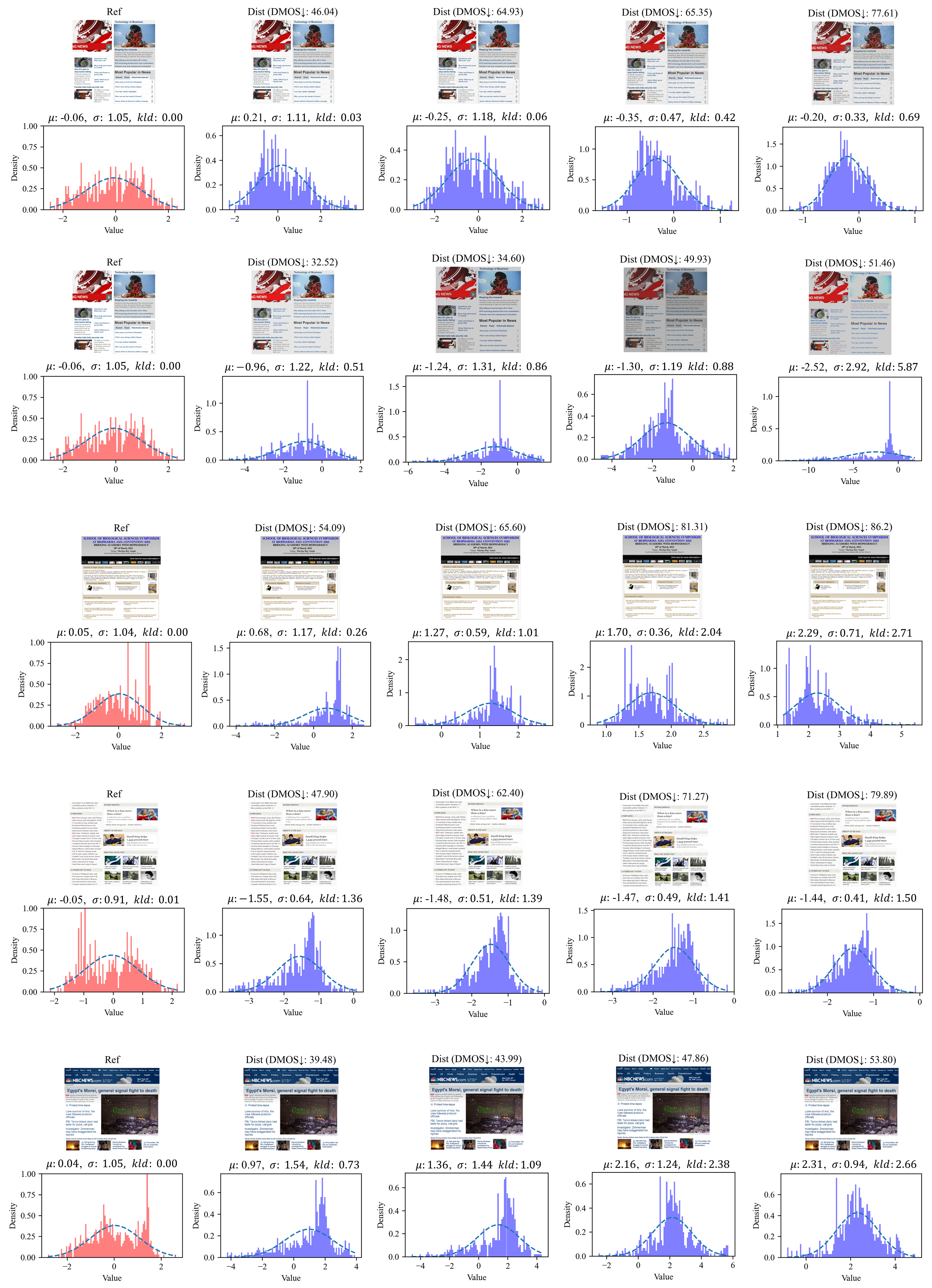}}
\caption{{\bl{Feature distribution visualization results. Sub-images in odd rows are the reference images and distorted images.   Sub-images in 
even rows are their corresponding feature distributions. In each row, the same dimension of the feature is selected for better comparison. From up to down, the distortion types are Gaussian Noise, Contrast Change, Gaussian Blur, JPEG2000, and Layer Segmentation based Coding. ``$\mu$'' and ``$\sigma$'' represent the mean and std of each feature distribution.  ``kld'' means the KL-divergence between the feature distribution and the standard Gaussian distribution. The ``$\downarrow$'' means the lower the value the better the quality.}}}
\label{fig:featv}
\end{figure*}

\begin{figure*}[!htb]
  \centerline{\includegraphics[width=0.9\linewidth]{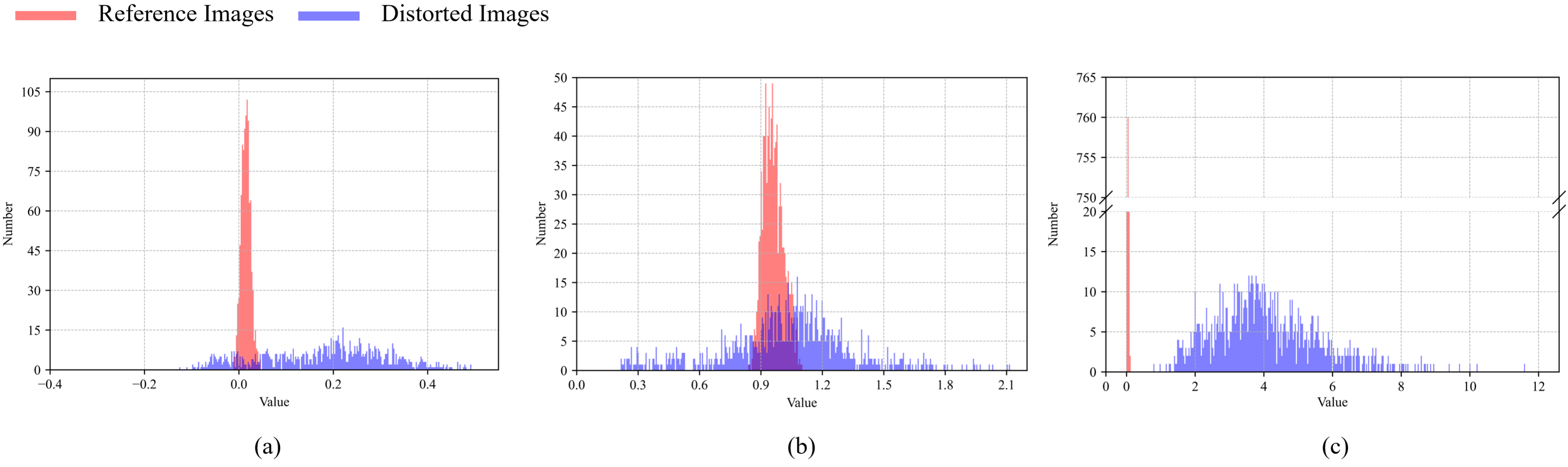}}
\caption{{\bl{Histograms of the mean values, std values, and KL divergences of the distortion-aware features extracted from reference images and distorted images. The KL divergences are measured between the feature distribution and the standard normal distribution. All images in the SIQAD dataset are presented. (a) Mean values. (b) std values. (c) KL divergence values.}}}
\label{fig:mskl}
\end{figure*}

\begin{table}[htbp]
  \centering
  \resizebox{3in}{!}{
  \caption{\bl{Performance on SIQAD dataset under the intra-dataset setting. The top two results are highlighted in boldface.}}
    \begin{tabular}{c|c|c|c}
    \toprule
    Method & PLCC  & SRCC  & RMSE \\
    \midrule
    NIQE~\cite{mittal2012making}  & 0.3415  & 0.3695  & 13.4670  \\
    BRISQUE~\cite{mittal2012no} & 0.7237  & 0.7708  & 8.2565  \\
    QAC~\cite{xue2013learning}   & 0.3751  & 0.3009  & 13.2690  \\
    IL-NIQE~\cite{zhang2015feature} & 0.3854  & 0.3575  & 13.9320  \\
    BQMS~\cite{gu2016learning}  & 0.7575  & 0.7251  & 9.3456  \\
    SIQE~\cite{gu2017no}  & 0.7906  & 0.7625  & 8.7650  \\
    ASIQE~\cite{gu2017no} & 0.7884  & 0.7570  & 8.8064  \\
    NRLT~\cite{fang2017no}  & 0.8442  & 0.8202  & 7.4156  \\
    HRFF~\cite{zheng2019no}  & 0.8520  & 0.8320  & 7.5960  \\
    CLGF~\cite{wu2019blind}  & 0.8331  & 0.8107  & 7.9172  \\
    DIQaM-NR~\cite{bosse2017deep}  & 0.8799  & 0.8662  & 6.7894  \\
    WaDIQaM-NR~\cite{bosse2017deep}& \textbf{0.8890 } & \textbf{0.8780 } & \textbf{6.7552 } \\
    Yang~\textit{et al.} (TIP21)~\cite{yang2021no} & 0.8529  & 0.8336  & 7.2817  \\
    Yang~\textit{et al.} (Tcy20)~\cite{yang2020no}& 0.8738  & 0.8543  & 6.9335  \\
    GraphIQA \cite{sun2022graphiqa} & 0.8493 & 0.8535 & 7.5770 \\
    VCRNet \cite{pan2022vcrnet} & 0.8544  & 0.8357  & 7.7165  \\
    \midrule
    DFSS-IQA (Ours) & \textbf{0.8818}  & \textbf{0.8820 } & \textbf{6.6684 } \\
    \bottomrule
    \end{tabular}%
    \label{tab:siqad}%
    }
  
\end{table}%

\subsection{Ablation Study}
In this subsection, to reveal the functionalities of different modules in our method, we perform the ablation studies on both SIQAD and SCID  databases with the cross-dataset settings. More specifically, three main modules are designed in our method including the Gaussian distribution regularization module, the semantic attention module, and the distortion type classification module, As shown in Table~\ref{tab:abl}, we denote the above three modules as  ``Gaus.", ``Attten.", and ``CLS", respectively. From the table, we can observe the worst performance is obtained when we ablate the attention module as the semantic-specific feature can adaptively adjust the relationship between the distribution divergence and the quality degradation. In addition, the performance drops again 
when we ablate the distortion type classification. This phenomenon demonstrates that distortion discrimination capability plays an important role in quality prediction. The Gaussian distribution regularization performed based on the MMD loss is ablated in the last experiment. In particular, we directly use the quality feature of $\boldsymbol{{F^{qua}}}$ for quality regression and distortion type classification. The ablation results on both the SIQAD and SCID datasets reveal the inferiority of the feature-based model when compared with the distribution-based one.  The best performance is achieved when all the modules are adopted, demonstrating the three modules contribute to the final performance. 

In Fig.~\ref{fig:lamda}, to explore the optimal   hyperparameter $\lambda_2$, we set its value to be 5e-1, 5e-2, 5e-3, 5e-4. We can observe the best performance is achieved when we set the parameter to 5e-3, and a large value or a smaller one will cause the performance to drop to some extent. The reason may lie in the trade-off between the strong distribution regularization and the variation in image quality.

\bbl{To explore an optimal output dimension of our multi-scale feature generator, we further conduct a study of the feature dimension setting in Table~\ref{tab:dm}. 
As shown in the table, a noticeable performance drop could be observed on the two cross-dataset settings (SCID $\rightarrow$ SIQAD and SIQAD $\rightarrow$ SCID) when the output dimension is reduced to 256. The reason may lie in the limited feature dimension which leads to the disentangled features can not fully capture the distortion-aware and semantic-aware information from the input image. Subsequently, we increase the feature dimension from 512 to 1792, and the performance improvement is only evidenced in the SCID $\rightarrow$ SIQAD setting, accompanied by an increase in model complexity. A significant performance drop can be observed when the dimension is increased to 2048, revealing a larger feature dimension could result in a negative effect on the model generalization capability.  Based upon the above observations, we set the final output dimension to 512, by which, the best trade-off between model accuracy and efficiency can be achieved.}

\begin{table}[htbp]
  \centering
  \resizebox{3in}{!}{
  \caption{\bl{Performance on SCID dataset under the intra-dataset setting. The top two results are highlighted in boldface.}}
    \begin{tabular}{c|c|c|c}
    \toprule
    Method & PLCC  & SRCC  & RMSE \\
    \midrule
    BLINDS-II~\cite{saad2012blind}  & 0.5851  & 0.5569  & 12.6253  \\
    NIQE~\cite{mittal2012making}  & 0.2931  & 0.2508  & 13.5401  \\
    BRISQUE~\cite{mittal2012no}  & 0.6004  & 0.5687  & 11 .6976 \\
    NFERM~\cite{gu2014using}  & 0.5928  & 0.5803  & 11.7647  \\
    IL-NIQE~\cite{zhang2015feature}  & 0.2573  & 0.2436  & 13.6852  \\
    BQMS~\cite{gu2016learning}  & 0.6188  & 0.6125  & 11.1251  \\
    SIQE~\cite{gu2017no}  & 0.6343  & 0.6009  & 10.9483  \\
    ASIQE~\cite{gu2017no}  & 0.6381  & 0.6046  & 10.5873  \\
    NRLT~\cite{fang2017no}  & 0.6216  & 0.6092  & 10.9042  \\
    CLGF~\cite{wu2019blind}  & 0.6978  & 0.6870  & 10.1439  \\
    DIQaM-NR~\cite{bosse2017deep} & 0.7086  & 0.6965  & 10.3085  \\
    WaDIQaM-NR~\cite{bosse2017deep} & 0.7885  & 0.7654  & 8.8189  \\
    Yang~\textit{et al.} (TIP21)~\cite{yang2021no} & 0.7147  & 0.6920  & 10.3988  \\
    Yang~\textit{et al.} (Tcy20)~\cite{yang2020no} & 0.7867  & 0.7562  & 8.5949  \\
    GraphIQA \cite{sun2022graphiqa} & \textbf{0.8342}  & \textbf{0.8309}  & \textbf{7.5717} \\
    VCRNet  \cite{pan2022vcrnet} & \textbf{0.8370}  & \textbf{0.8332}  & 9.1558 \\
    \midrule
    DFSS-IQA (Ours) & 0.8138  & 0.8146  & \textbf{8.0125 } \\
    \bottomrule
    \end{tabular}%
    \label{tab:scid}%
    }
  
\end{table}%

\begin{table}[htbp]
  \centering
  \caption{Ablation studies under two cross-dataset settings.}
  \resizebox{3in}{!}{
    \begin{tabular}{c|c|c|cc|cc}
    \toprule
    \multicolumn{3}{c|}{Setting} & \multicolumn{2}{c|}{SIQAD $\rightarrow$ SCID} & \multicolumn{2}{c}{SCID $\rightarrow$ SIQAD} \\
    \midrule
    Gaus.  & Atten. & CLS   & SRCC  & PLCC  & SRCC  & PLCC \\
    \midrule
    \XSolidBrush      & \XSolidBrush     & \Checkmark     & 0.7570 & 0.7697 & 0.7966 & 0.8174 \\
    \Checkmark    & \XSolidBrush      & \Checkmark    & 0.7159 & 0.7217 & 0.7446 & 0.7732 \\
    \Checkmark     & \Checkmark     & \XSolidBrush      & 0.7210 & 0.7275 & 0.7483 & 0.7841 \\
    \Checkmark     & \Checkmark     & \Checkmark     & \textbf{0.7669 } & \textbf{0.7815 } & \textbf{0.7969 } & \textbf{0.8344 } \\
    \bottomrule
    \end{tabular}%
    }
  \label{tab:abl}%
\end{table}%

\begin{table*}[t]
  \centering
  \caption{\bbl{Study results of different output feature dimensions. The ``SCID $\rightarrow$ SIQAD'' (or ``SIQAD $\rightarrow$ SCID'') indicates that the model is trained on the SCID (or SIQAD) dataset and subsequently tested on the SIQAD (or SCID) dataset.  `` \bbl{$\uparrow$}'' denotes an increase in value compared to the original dimension setting (512), and ``\br{$\downarrow$}'' represents a declining value. The FLOPS is estimated with one image patch as input.}}
  \resizebox{7.0in}{!}{
    \begin{tabular}{c|cc|cc|cc|cc}
    \toprule
    \multicolumn{1}{l|}{Dimension} & \multicolumn{2}{c|}{256} & \multicolumn{2}{c|}{512 (Original)} & \multicolumn{2}{c|}{1792} & \multicolumn{2}{c}{2048} \\
    \midrule
    \multicolumn{1}{l|}{Setting} & \multicolumn{1}{c|}{ SCID $\rightarrow$ SIQAD} &  SIQAD  $\rightarrow$ SCID & \multicolumn{1}{c|}{ SCID $\rightarrow$ SIQAD} &  SIQAD  $\rightarrow$ SCID & \multicolumn{1}{c|}{ SCID $\rightarrow$ SIQAD} &  SIQAD $\rightarrow$  SCID & \multicolumn{1}{c|}{ SCID  $\rightarrow$ SIQAD} &  SIQAD  $\rightarrow$ SCID \\
    \midrule
    \multicolumn{1}{l|}{SRCC}  & \multicolumn{1}{c|}{0.7956 \br{$\downarrow$}} & 0.7310 \br{$\downarrow$} & \multicolumn{1}{c|}{0.7969} & 0.7669 & \multicolumn{1}{c|}{0.8119 \bbl{$\uparrow$}} & 0.7512 \br{$\downarrow$} & \multicolumn{1}{c|}{0.7911 \br{$\downarrow$}} & 0.7427 \br{$\downarrow$} \\
    \midrule
   \multicolumn{1}{l|}{FLOPS (G)}  & \multicolumn{2}{c|}{0.4583 \br{$\downarrow$}} & \multicolumn{2}{c|}{0.4585} & \multicolumn{2}{c|}{0.4592 \bbl{$\uparrow$}} & \multicolumn{2}{c}{0.4594 \bbl{$\uparrow$}} \\
    \bottomrule
    \end{tabular}%
    }
  \label{tab:dm}%
\end{table*}%

\subsection{Feature Visualization}
\bl{In our method, we disentangle the quality-aware feature into a semantic-specific feature and a distortion-aware feature. To verify the disentanglement, a t-SNE~\cite{maaten2008visualizing} visualization is adopted in Fig.~\ref{fig:tsne}. In particular, the two types of features extracted from images of the SIQAD dataset are visualized. From the figure, we can observe: 1) The distortion-aware features of the reference images are clustered into one group as their distortions are the same though the images contain different contents. In comparison, the features of distorted images are distributed in different spaces by their distortion levels, revealing the features are of high awareness of image distortions. 2) Regarding the semantic-specific feature, the images are clustered into one group when they are with the same content and the total number of content clusters is 20 which corresponds to the number of contents in the SIQAD dataset. This phenomenon reveals that the disentangled features are of high awareness of image semantics.}

\bl{To better understand the performance of our method, we visualize the quality relevant features  $\boldsymbol{F^{rd}}$ and  $\boldsymbol{F^{dd}}$ in  Fig.~\ref{fig:featv}.  More specifically, the model learned on the SIQAD dataset is adopted for feature extraction, and for each row, we randomly select one dimension of $\boldsymbol{F^{rd}}$ and  $\boldsymbol{F^{dd}}$ for their distribution visualization. As shown in  Fig.~\ref{fig:featv}, the distributions of $\boldsymbol{F^{rd}}$ and  $\boldsymbol{F^{ad}}$ present different mean and std values due to the standard distribution being corrupted by the imposed distortion. In particular, the KL divergence increases as the distortion becomes severe, revealing that the distribution divergence is able to reflect the distortion levels effectively. In addition, we can observe that distortion-awareness exists in different dimensions of the distortion-aware features, revealing that the quality prediction should be an aggregation of the distributions at different dimensions. In Fig.~\ref{fig:mskl}, we further present the histograms of the mean values, std values, and KL divergences of the distortion-aware features, from which, we can observe the mean values of the feature distributions of the pristine images are close to 0, std values are close to 1, and the KL divergences are significantly small, demonstrating a unified distribution that the pristine SCIs obey. In comparison, the mean values, std values, and KL divergences of the distorted images are dispersed, revealing the feature statistics would be destructed due to the distortions.}

\section{Conclusions}
In this paper, we propose a novel NR-SCIQA method by constructing the specific statistics of SCIs in the deep feature space.  In particular, we first extract the quality-aware feature at multi-scales and disentangle it into the distortion-aware feature and semantic-specific feature. The unified distribution constraint is imposed on the distortion-aware feature, aiming for the construction of the statistics of SCI. Finally, the image quality can be estimated by combining both the distribution information and semantic information. Experimental results have demonstrated the superior generalization capability of our scene statistic-based model, especially in cross-dataset settings. The proposed method sheds light on the exploration of the intrinsic statistics of SCIs and provides potential guidance for high-quality image generation with computers.


\bibliographystyle{IEEEtran}
\bibliography{DFSS}

\begin{thebibliography}{10}
\providecommand{\url}[1]{#1}
\csname url@samestyle\endcsname
\providecommand{\newblock}{\relax}
\providecommand{\bibinfo}[2]{#2}
\providecommand{\BIBentrySTDinterwordspacing}{\spaceskip=0pt\relax}
\providecommand{\BIBentryALTinterwordstretchfactor}{4}
\providecommand{\BIBentryALTinterwordspacing}{\spaceskip=\fontdimen2\font plus
\BIBentryALTinterwordstretchfactor\fontdimen3\font minus \fontdimen4\font\relax}
\providecommand{\BIBforeignlanguage}[2]{{%
\expandafter\ifx\csname l@#1\endcsname\relax
\typeout{** WARNING: IEEEtran.bst: No hyphenation pattern has been}%
\typeout{** loaded for the language `#1'. Using the pattern for}%
\typeout{** the default language instead.}%
\else
\language=\csname l@#1\endcsname
\fi
#2}}
\providecommand{\BIBdecl}{\relax}
\BIBdecl

\bibitem{wang2016just}
S.~Wang, L.~Ma, Y.~Fang, W.~Lin, S.~Ma, and W.~Gao, ``Just noticeable difference estimation for screen content images,'' \emph{IEEE Transactions on Image Processing}, vol.~25, no.~8, pp. 3838--3851, 2016.

\bibitem{wang2016reduced}
S.~Wang, K.~Gu, X.~Zhang, W.~Lin, S.~Ma, and W.~Gao, ``Reduced-reference quality assessment of screen content images,'' \emph{IEEE Transactions on Circuits and Systems for Video Technology}, vol.~28, no.~1, pp. 1--14, 2016.

\bibitem{wang2015joint}
S.~Wang, K.~Gu, S.~Ma, and W.~Gao, ``Joint chroma downsampling and upsampling for screen content image,'' \emph{IEEE Transactions on Circuits and Systems for Video Technology}, vol.~26, no.~9, pp. 1595--1609, 2015.

\bibitem{xu2016parsimonious}
Q.~Xu, J.~Xiong, X.~Cao, and Y.~Yao, ``Parsimonious mixed-effects hodgerank for crowdsourced preference aggregation,'' in \emph{ACM International Conference on Multimedia (ACM MM)}, 2016, pp. 841--850.

\bibitem{xu2013online}
Q.~Xu, J.~Xiong, Q.~Huang, and Y.~Yao, ``Online hodgerank on random graphs for crowdsourceable {QoE} evaluation,'' \emph{IEEE Transactions on Multimedia}, vol.~16, no.~2, pp. 373--386, 2013.

\bibitem{xu2013robust}
{Q. Xu, J. Xiong, Q. Huang, and Y. Yao}, ``Robust evaluation for quality of experience in crowdsourcing,'' in \emph{ACM International Conference Multimedia (ACM MM)}, 2013, pp. 43--52.

\bibitem{xu2012online}
Q.~Xu, Q.~Huang, and Y.~Yao, ``Online crowdsourcing subjective image quality assessment,'' in \emph{ACM International Conference Multimedia (ACM MM)}, 2012, pp. 359--368.

\bibitem{xu2012hodgerank}
Q.~Xu, Q.~Huang, T.~Jiang, B.~Yan, W.~Lin, and Y.~Yao, ``{HodgeRank} on random graphs for subjective video quality assessment,'' \emph{IEEE Transactions on Multimedia}, vol.~14, no.~3, pp. 844--857, 2012.

\bibitem{min2021screen}
X.~Min, K.~Gu, G.~Zhai, X.~Yang, W.~Zhang, P.~Le~Callet, and C.~W. Chen, ``Screen content quality assessment: overview, benchmark, and beyond,'' \emph{ACM Computing Surveys}, vol.~54, no.~9, pp. 1--36, 2021.

\bibitem{fang2017objective}
Y.~Fang, J.~Yan, J.~Liu, S.~Wang, Q.~Li, and Z.~Guo, ``Objective quality assessment of screen content images by uncertainty weighting,'' \emph{IEEE Transactions on Image Processing}, vol.~26, no.~4, pp. 2016--2027, 2017.

\bibitem{ni2017esim}
Z.~Ni, L.~Ma, H.~Zeng, J.~Chen, C.~Cai, and K.-K. Ma, ``{ESIM}: Edge similarity for screen content image quality assessment,'' \emph{IEEE Transactions on Image Processing}, vol.~26, no.~10, pp. 4818--4831, 2017.

\bibitem{ni2018gabor}
Z.~Ni, H.~Zeng, L.~Ma, J.~Hou, J.~Chen, and K.-K. Ma, ``A gabor feature-based quality assessment model for the screen content images,'' \emph{IEEE Transactions on Image Processing}, vol.~27, no.~9, pp. 4516--4528, 2018.

\bibitem{zhang2018quality}
Y.~Zhang, D.~M. Chandler, and X.~Mou, ``Quality assessment of screen content images via convolutional-neural-network-based synthetic/natural segmentation,'' \emph{IEEE Transactions on Image Processing}, vol.~27, no.~10, pp. 5113--5128, 2018.

\bibitem{yang2021full}
J.~Yang, Z.~Bian, Y.~Zhao, W.~Lu, and X.~Gao, ``Full-reference quality assessment for screen content images based on the concept of global-guidance and local-adjustment,'' \emph{IEEE Transactions on Broadcasting}, vol.~67, no.~3, pp. 696--709, 2021.

\bibitem{gu2017no}
K.~Gu, J.~Zhou, J.-F. Qiao, G.~Zhai, W.~Lin, and A.~C. Bovik, ``No-reference quality assessment of screen content pictures,'' \emph{IEEE Transactions on Image Processing}, vol.~26, no.~8, pp. 4005--4018, 2017.

\bibitem{fang2017no}
Y.~Fang, J.~Yan, L.~Li, J.~Wu, and W.~Lin, ``No reference quality assessment for screen content images with both local and global feature representation,'' \emph{IEEE Transactions on Image Processing}, vol.~27, no.~4, pp. 1600--1610, 2017.

\bibitem{zheng2019no}
L.~Zheng, L.~Shen, J.~Chen, P.~An, and J.~Luo, ``No-reference quality assessment for screen content images based on hybrid region features fusion,'' \emph{IEEE Transactions on Multimedia}, vol.~21, no.~8, pp. 2057--2070, 2019.

\bibitem{jiang2019deep}
X.~Jiang, L.~Shen, G.~Feng, L.~Yu, and P.~An, ``Deep optimization model for screen content image quality assessment using neural networks,'' \emph{arXiv preprint arXiv:1903.00705}, 2019.

\bibitem{cheng2018fast}
Z.~Cheng, M.~Takeuchi, K.~Kanai, and J.~Katto, ``A fast no-reference screen content image quality prediction using convolutional neural networks,'' in \emph{IEEE International Conference on Multimedia \& Expo Workshops (ICMEW)}, 2018, pp. 1--6.

\bibitem{chen2018naturalization}
J.~Chen, L.~Shen, L.~Zheng, and X.~Jiang, ``Naturalization module in neural networks for screen content image quality assessment,'' \emph{IEEE Signal Processing Letters}, vol.~25, no.~11, pp. 1685--1689, 2018.

\bibitem{chen2021no}
B.~Chen, H.~Li, H.~Fan, and S.~Wang, ``No-reference screen content image quality assessment with unsupervised domain adaptation,'' \emph{IEEE Transactions on Image Processing}, vol.~30, pp. 5463--5476, 2021.

\bibitem{yang2021staged}
J.~Yang, Z.~Bian, Y.~Zhao, W.~Lu, and X.~Gao, ``Staged-learning: Assessing the quality of screen content images from distortion information,'' \emph{IEEE Signal Processing Letters}, vol.~28, pp. 1480--1484, 2021.

\bibitem{ponomarenko2015image}
N.~Ponomarenko, L.~Jin, O.~Ieremeiev, V.~Lukin, K.~Egiazarian, J.~Astola, B.~Vozel, K.~Chehdi, M.~Carli, F.~Battisti \emph{et~al.}, ``Image database {TID2013}: {P}eculiarities, results and perspectives,'' \emph{Signal Processing: Image Communication}, vol.~30, pp. 57--77, 2015.

\bibitem{yang2015perceptual}
H.~Yang, Y.~Fang, and W.~Lin, ``Perceptual quality assessment of screen content images,'' \emph{IEEE Transactions on Image Processing}, vol.~24, no.~11, pp. 4408--4421, 2015.

\bibitem{mittal2012making}
A.~Mittal, R.~Soundararajan, and A.~C. Bovik, ``Making a “completely blind” image quality analyzer,'' \emph{IEEE Signal Processing Letters}, vol.~20, no.~3, pp. 209--212, 2012.

\bibitem{mittal2012no}
A.~Mittal, A.~K. Moorthy, and A.~C. Bovik, ``No-reference image quality assessment in the spatial domain,'' \emph{IEEE Transactions on Image Processing}, vol.~21, no.~12, pp. 4695--4708, 2012.

\bibitem{moorthy2010two}
A.~K. Moorthy and A.~C. Bovik, ``A two-step framework for constructing blind image quality indices,'' \emph{IEEE Signal Processing Letters}, vol.~17, no.~5, pp. 513--516, 2010.

\bibitem{moorthy2011blind}
{Moorthy, Anush Krishna and Bovik, Alan Conrad}, ``Blind image quality assessment: From natural scene statistics to perceptual quality,'' \emph{IEEE Transactions on Image Processing}, vol.~20, no.~12, pp. 3350--3364, 2011.

\bibitem{gu2016saliency}
K.~Gu, S.~Wang, H.~Yang, W.~Lin, G.~Zhai, X.~Yang, and W.~Zhang, ``Saliency-guided quality assessment of screen content images,'' \emph{IEEE Transactions on Multimedia}, vol.~18, no.~6, pp. 1098--1110, 2016.

\bibitem{ni2017scid}
Z.~Ni, L.~Ma, H.~Zeng, Y.~Fu, L.~Xing, and K.-K. Ma, ``{SCID}: A database for screen content images quality assessment,'' in \emph{IEEE International Symposium on Intelligent Signal Processing and Communication Systems (ISPACS)}, 2017, pp. 774--779.

\bibitem{tang2011learning}
H.~Tang, N.~Joshi, and A.~Kapoor, ``Learning a blind measure of perceptual image quality,'' in \emph{IEEE Conference on Computer Vision and Pattern Recognition (CVPR)}, 2011, pp. 305--312.

\bibitem{wu2015blind}
Q.~Wu, H.~Li, F.~Meng, K.~N. Ngan, B.~Luo, C.~Huang, and B.~Zeng, ``Blind image quality assessment based on multichannel feature fusion and label transfer,'' \emph{IEEE Transactions on Circuits and Systems for Video Technology}, vol.~26, no.~3, pp. 425--440, 2015.

\bibitem{friston2006free}
K.~Friston, J.~Kilner, and L.~Harrison, ``A free energy principle for the brain,'' \emph{Journal of Physiology-Paris}, vol. 100, no. 1-3, pp. 70--87, 2006.

\bibitem{friston2010free}
K.~Friston, ``The free-energy principle: a unified brain theory?'' \emph{Nature Reviews Neuroscience}, vol.~11, no.~2, pp. 127--138, 2010.

\bibitem{gu2013no}
K.~Gu, G.~Zhai, X.~Yang, W.~Zhang, and L.~Liang, ``No-reference image quality assessment metric by combining free energy theory and structural degradation model,'' in \emph{IEEE International Conference on Multimedia and Expo (ICME)}, 2013, pp. 1--6.

\bibitem{gu2014using}
K.~Gu, G.~Zhai, X.~Yang, and W.~Zhang, ``Using free energy principle for blind image quality assessment,'' \emph{IEEE Transactions on Multimedia}, vol.~17, no.~1, pp. 50--63, 2014.

\bibitem{zhai2011psychovisual}
G.~Zhai, X.~Wu, X.~Yang, W.~Lin, and W.~Zhang, ``A psychovisual quality metric in free-energy principle,'' \emph{IEEE Transactions on Image Processing}, vol.~21, no.~1, pp. 41--52, 2011.

\bibitem{xue2014blind}
W.~Xue, X.~Mou, L.~Zhang, A.~C. Bovik, and X.~Feng, ``Blind image quality assessment using joint statistics of gradient magnitude and laplacian features,'' \emph{IEEE Transactions on Image Processing}, vol.~23, no.~11, pp. 4850--4862, 2014.

\bibitem{ye2012unsupervised}
P.~Ye, J.~Kumar, L.~Kang, and D.~Doermann, ``Unsupervised feature learning framework for no-reference image quality assessment,'' in \emph{IEEE Conference on Computer Vision and Pattern Recognition (CVPR)}, 2012, pp. 1098--1105.

\bibitem{ye2012no}
P.~Ye and D.~Doermann, ``No-reference image quality assessment using visual codebooks,'' \emph{IEEE Transactions on Image Processing}, vol.~21, no.~7, pp. 3129--3138, 2012.

\bibitem{zhang2015som}
P.~Zhang, W.~Zhou, L.~Wu, and H.~Li, ``{SOM}: Semantic obviousness metric for image quality assessment,'' in \emph{IEEE Conference on Computer Vision and Pattern Recognition (CVPR)}, 2015, pp. 2394--2402.

\bibitem{kang2014convolutional}
L.~Kang, P.~Ye, Y.~Li, and D.~Doermann, ``Convolutional neural networks for no-reference image quality assessment,'' in \emph{IEEE Conference on Computer Vision and Pattern Recognition (CVPR)}, 2014, pp. 1733--1740.

\bibitem{bianco2018use}
S.~Bianco, L.~Celona, P.~Napoletano, and R.~Schettini, ``On the use of deep learning for blind image quality assessment,'' \emph{Signal, Image and Video Processing}, vol.~12, no.~2, pp. 355--362, 2018.

\bibitem{kang2015simultaneous}
L.~Kang, P.~Ye, Y.~Li, and D.~Doermann, ``Simultaneous estimation of image quality and distortion via multi-task convolutional neural networks,'' in \emph{IEEE International Conference on Image Processing (ICIP)}, 2015, pp. 2791--2795.

\bibitem{liu2017rankiqa}
X.~Liu, J.~van~de Weijer, and A.~D. Bagdanov, ``{RankIQA}: Learning from rankings for no-reference image quality assessment,'' in \emph{IEEE International Conference on Computer Vision (ICCV)}, 2017, pp. 1040--1049.

\bibitem{niu2019siamese}
Y.~Niu, D.~Huang, Y.~Shi, and X.~Ke, ``Siamese-network-based learning to rank for no-reference {2D} and {3D} image quality assessment,'' \emph{IEEE Access}, vol.~7, pp. 101\,583--101\,595, 2019.

\bibitem{ying2020quality}
Z.~Ying, D.~Pan, and P.~Shi, ``Quality difference ranking model for smartphone camera photo quality assessment,'' in \emph{International Conference on Multimedia \& Expo Workshops (ICMEW)}, 2020, pp. 1--6.

\bibitem{zhang2018blind}
W.~Zhang, K.~Ma, J.~Yan, D.~Deng, and Z.~Wang, ``Blind image quality assessment using a deep bilinear convolutional neural network,'' \emph{IEEE Transactions on Circuits and Systems for Video Technology}, vol.~30, no.~1, pp. 36--47, 2018.

\bibitem{ma2017end}
K.~Ma, W.~Liu, K.~Zhang, Z.~Duanmu, Z.~Wang, and W.~Zuo, ``End-to-end blind image quality assessment using deep neural networks,'' \emph{IEEE Transactions on Image Processing}, vol.~27, no.~3, pp. 1202--1213, 2017.

\bibitem{fang2020perceptual}
Y.~Fang, H.~Zhu, Y.~Zeng, K.~Ma, and Z.~Wang, ``Perceptual quality assessment of smartphone photography,'' in \emph{IEEE Conference on Computer Vision and Pattern Recognition (CVPR)}, 2020, pp. 3677--3686.

\bibitem{ma2023forgetting}
R.~Ma, Q.~Wu, K.~N. Ngan, H.~Li, F.~Meng, and L.~Xu, ``Forgetting to remember: A scalable incremental learning framework for cross-task blind image quality assessment,'' \emph{IEEE Transactions on Multimedia}, 2023.

\bibitem{ma2021remember}
R.~Ma, H.~Luo, Q.~Wu, K.~N. Ngan, H.~Li, F.~Meng, and L.~Xu, ``Remember and reuse: Cross-task blind image quality assessment via relevance-aware incremental learning,'' in \emph{ACM International Conference on Multimedia (ACM MM)}, 2021, pp. 5248--5256.

\bibitem{chen2022spiq}
P.~Chen, L.~Li, Q.~Wu, and J.~Wu, ``{SPIQ}: A self-supervised pre-trained model for image quality assessment,'' \emph{IEEE Signal Processing Letters}, vol.~29, pp. 513--517, 2022.

\bibitem{yue2022semi}
G.~Yue, D.~Cheng, L.~Li, T.~Zhou, H.~Liu, and T.~Wang, ``Semi-supervised authentically distorted image quality assessment with consistency-preserving dual-branch convolutional neural network,'' \emph{IEEE Transactions on Multimedia}, vol.~25, pp. 6499--6511, 2022.

\bibitem{lin2018hallucinated}
K.-Y. Lin and G.~Wang, ``{Hallucinated-IQA}: No-reference image quality assessment via adversarial learning,'' in \emph{IEEE Conference on Computer Vision and Pattern Recognition (CVPR)}, 2018, pp. 732--741.

\bibitem{chen2022no}
B.~Chen, L.~Zhu, C.~Kong, H.~Zhu, S.~Wang, and Z.~Li, ``No-reference image quality assessment by hallucinating pristine features,'' \emph{IEEE Transactions on Image Processing}, vol.~31, pp. 6139--6151, 2022.

\bibitem{yue2019blind}
G.~Yue, C.~Hou, W.~Yan, L.~K. Choi, T.~Zhou, and Y.~Hou, ``Blind quality assessment for screen content images via convolutional neural network,'' \emph{Digital Signal Processing}, vol.~91, pp. 21--30, 2019.

\bibitem{jiang2020no}
X.~Jiang, L.~Shen, L.~Yu, M.~Jiang, and G.~Feng, ``No-reference screen content image quality assessment based on multi-region features,'' \emph{Neurocomputing}, vol. 386, pp. 30--41, 2020.

\bibitem{min2018saliency}
X.~Min, K.~Gu, G.~Zhai, M.~Hu, and X.~Yang, ``Saliency-induced reduced-reference quality index for natural scene and screen content images,'' \emph{Signal Processing}, vol. 145, pp. 127--136, 2018.

\bibitem{min2017unified}
X.~Min, K.~Ma, K.~Gu, G.~Zhai, Z.~Wang, and W.~Lin, ``Unified blind quality assessment of compressed natural, graphic, and screen content images,'' \emph{IEEE Transactions on Image Processing}, vol.~26, no.~11, pp. 5462--5474, 2017.

\bibitem{li2017exploiting}
D.~Li, T.~Jiang, and M.~Jiang, ``Exploiting high-level semantics for no-reference image quality assessment of realistic blur images,'' in \emph{ACM International Conference on Multimedia (ACM MM)}, 2017, pp. 378--386.

\bibitem{li2018has}
D.~Li, T.~Jiang, W.~Lin, and M.~Jiang, ``Which has better visual quality: The clear blue sky or a blurry animal?'' \emph{IEEE Transactions on Multimedia}, vol.~21, no.~5, pp. 1221--1234, 2018.

\bibitem{zhang2018unreasonable}
R.~Zhang, P.~Isola, A.~A. Efros, E.~Shechtman, and O.~Wang, ``The unreasonable effectiveness of deep features as a perceptual metric,'' in \emph{IEEE Conference on Computer Vision and Pattern Recognition (CVPR)}, 2018, pp. 586--595.

\bibitem{bosse2017deep}
S.~Bosse, D.~Maniry, K.-R. M{\"u}ller, T.~Wiegand, and W.~Samek, ``Deep neural networks for no-reference and full-reference image quality assessment,'' \emph{IEEE Transactions on Image Processing}, vol.~27, no.~1, pp. 206--219, 2017.

\bibitem{ding2020image}
K.~Ding, K.~Ma, S.~Wang, and E.~P. Simoncelli, ``Image quality assessment: {U}nifying structure and texture similarity,'' \emph{IEEE Transactions on Pattern Analysis and Machine Intelligence}, vol.~44, no.~5, pp. 2567--2581, 2020.

\bibitem{he2015spatial}
K.~He, X.~Zhang, S.~Ren, and J.~Sun, ``Spatial pyramid pooling in deep convolutional networks for visual recognition,'' \emph{IEEE Transactions on Pattern Analysis and Machine Intelligence}, vol.~37, no.~9, pp. 1904--1916, 2015.

\bibitem{kim2016fully}
J.~Kim and S.~Lee, ``Fully deep blind image quality predictor,'' \emph{IEEE Journal of Selected Topics in Signal Processing}, vol.~11, no.~1, pp. 206--220, 2016.

\bibitem{chen2021learning}
B.~Chen, L.~Zhu, G.~Li, F.~Lu, H.~Fan, and S.~Wang, ``Learning generalized spatial-temporal deep feature representation for no-reference video quality assessment,'' \emph{IEEE Transactions on Circuits and Systems for Video Technology}, vol.~32, no.~4, pp. 1903--1916, 2021.

\bibitem{zhu2022learning}
H.~Zhu, B.~Chen, L.~Zhu, and S.~Wang, ``Learning spatiotemporal interactions for user-generated video quality assessment,'' \emph{IEEE Transactions on Circuits and Systems for Video Technology}, vol.~33, no.~3, pp. 1031--1042, 2022.

\bibitem{gretton2012kernel}
A.~Gretton, K.~M. Borgwardt, M.~J. Rasch, B.~Sch{\"o}lkopf, and A.~Smola, ``A kernel two-sample test,'' \emph{Journal of Machine Learning Research}, vol.~13, no. Mar, pp. 723--773, 2012.

\bibitem{paszke2019pytorch}
A.~Paszke, S.~Gross, F.~Massa, A.~Lerer, J.~Bradbury, G.~Chanan, T.~Killeen, Z.~Lin, N.~Gimelshein, L.~Antiga \emph{et~al.}, ``{PyTorch}: An imperative style, high-performance deep learning library,'' in \emph{Advances in Neural Information Processing Systems}, 2019, pp. 8026--8037.

\bibitem{kingma2014adam}
D.~P. Kingma and J.~Ba, ``Adam: A method for stochastic optimization,'' \emph{arXiv preprint arXiv:1412.6980}, 2014.

\bibitem{zhang2015feature}
L.~Zhang, L.~Zhang, and A.~C. Bovik, ``A feature-enriched completely blind image quality evaluator,'' \emph{IEEE Transactions on Image Processing}, vol.~24, no.~8, pp. 2579--2591, 2015.

\bibitem{xu2016blind}
J.~Xu, P.~Ye, Q.~Li, H.~Du, Y.~Liu, and D.~Doermann, ``Blind image quality assessment based on high order statistics aggregation,'' \emph{IEEE Transactions on Image Processing}, vol.~25, no.~9, pp. 4444--4457, 2016.

\bibitem{gu2016learning}
K.~Gu, G.~Zhai, W.~Lin, X.~Yang, and W.~Zhang, ``Learning a blind quality evaluation engine of screen content images,'' \emph{Neurocomputing}, vol. 196, pp. 140--149, 2016.

\bibitem{fang2019perceptual}
Y.~Fang, R.~Du, Y.~Zuo, W.~Wen, and L.~Li, ``Perceptual quality assessment for screen content images by spatial continuity,'' \emph{IEEE Transactions on Circuits and Systems for Video Technology}, vol.~30, no.~11, pp. 4050--4063, 2019.

\bibitem{sun2022graphiqa}
S.~Sun, T.~Yu, J.~Xu, W.~Zhou, and Z.~Chen, ``{GraphIQA}: Learning distortion graph representations for blind image quality assessment,'' \emph{IEEE Transactions on Multimedia}, 2022.

\bibitem{pan2022vcrnet}
Z.~Pan, F.~Yuan, J.~Lei, Y.~Fang, X.~Shao, and S.~Kwong, ``{VCRNet}: Visual compensation restoration network for no-reference image quality assessment,'' \emph{IEEE Transactions on Image Processing}, vol.~31, pp. 1613--1627, 2022.

\bibitem{xue2013learning}
W.~Xue, L.~Zhang, and X.~Mou, ``Learning without human scores for blind image quality assessment,'' in \emph{IEEE Conference on Computer Vision and Pattern Recognition (CVPR)}, 2013, pp. 995--1002.

\bibitem{wu2019blind}
J.~Wu, Z.~Xia, H.~Zhang, and H.~Li, ``Blind quality assessment for screen content images by combining local and global features,'' \emph{Digital Signal Processing}, vol.~91, pp. 31--40, 2019.

\bibitem{li2019cnn}
R.~Li, H.~Yang, T.~Yu, and Z.~Pan, ``{CNN} model for screen content image quality assessment based on region difference,'' in \emph{International Conference on Signal and Image Processing (ICSIP)}, 2019, pp. 1010--1014.

\bibitem{yang2020no}
J.~Yang, Y.~Zhao, J.~Liu, B.~Jiang, Q.~Meng, W.~Lu, and X.~Gao, ``No reference quality assessment for screen content images using stacked autoencoders in pictorial and textual regions,'' \emph{IEEE Transactions on Cybernetics}, vol.~52, no.~5, pp. 2798--2810, 2020.

\bibitem{yang2021no}
J.~Yang, Z.~Bian, J.~Liu, B.~Jiang, W.~Lu, X.~Gao, and H.~Song, ``No-reference quality assessment for screen content images using visual edge model and adaboosting neural network,'' \emph{IEEE Transactions on Image Processing}, vol.~30, pp. 6801--6814, 2021.

\bibitem{2019kadid}
H.~Lin, V.~Hosu, and D.~Saupe, ``{KADID-10k}: A large-scale artificially distorted {IQA} database,'' in \emph{International Conference on Quality of Multimedia Experience (QoMEX)}, 2019.

\bibitem{ma2016waterloo}
K.~Ma, Z.~Duanmu, Q.~Wu, Z.~Wang, H.~Yong, H.~Li, and L.~Zhang, ``Waterloo exploration database: New challenges for image quality assessment models,'' \emph{IEEE Transactions on Image Processing}, vol.~26, no.~2, pp. 1004--1016, 2016.

\bibitem{deng2009imagenet}
J.~Deng, W.~Dong, R.~Socher, L.-J. Li, K.~Li, and L.~Fei-Fei, ``Image{N}et: {A} large-scale hierarchical image database,'' in \emph{IEEE Conference on Computer Vision and Pattern Recognition}, 2009, pp. 248--255.

\bibitem{maaten2008visualizing}
L.~v.~d. Maaten and G.~Hinton, ``Visualizing data using t-{SNE},'' \emph{Journal of Machine Learning Research}, vol.~9, no. Nov, pp. 2579--2605, 2008.

\bibitem{saad2012blind}
M.~A. Saad, A.~C. Bovik, and C.~Charrier, ``Blind image quality assessment: A natural scene statistics approach in the dct domain,'' \emph{IEEE Transactions on Image Processing}, vol.~21, no.~8, pp. 3339--3352, 2012.

\end{thebibliography}


\begin{IEEEbiography}[{\includegraphics[width=1in,height=1.25in, clip,keepaspectratio]{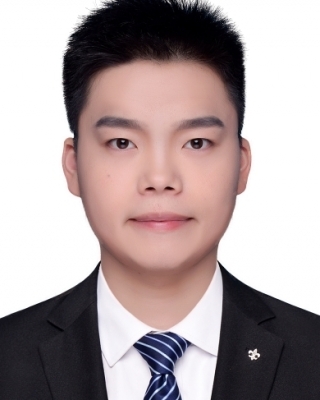}}]{Baoliang Chen} (Member, IEEE) received his B.E. degree in Electronic Information Science and Technology from Hefei University of Technology, Hefei, China, in 2015, his M.S. degree in Intelligent Information Processing from Xidian University, Xian, China, in 2018, and his Ph.D. degree in computer science from the City University of Hong Kong, Hong Kong, in 2022.  From 2022 to 2024, he was a postdoctoral researcher with the Department of Computer Science, City University of Hong Kong. He is currently an Associate Professor with the Department of Computer Science, South China Normal University.  His research interests include image/video quality assessment and transfer learning.
\end{IEEEbiography}

\begin{IEEEbiography}[{\includegraphics[width=1in,height=1.25in,clip,keepaspectratio]{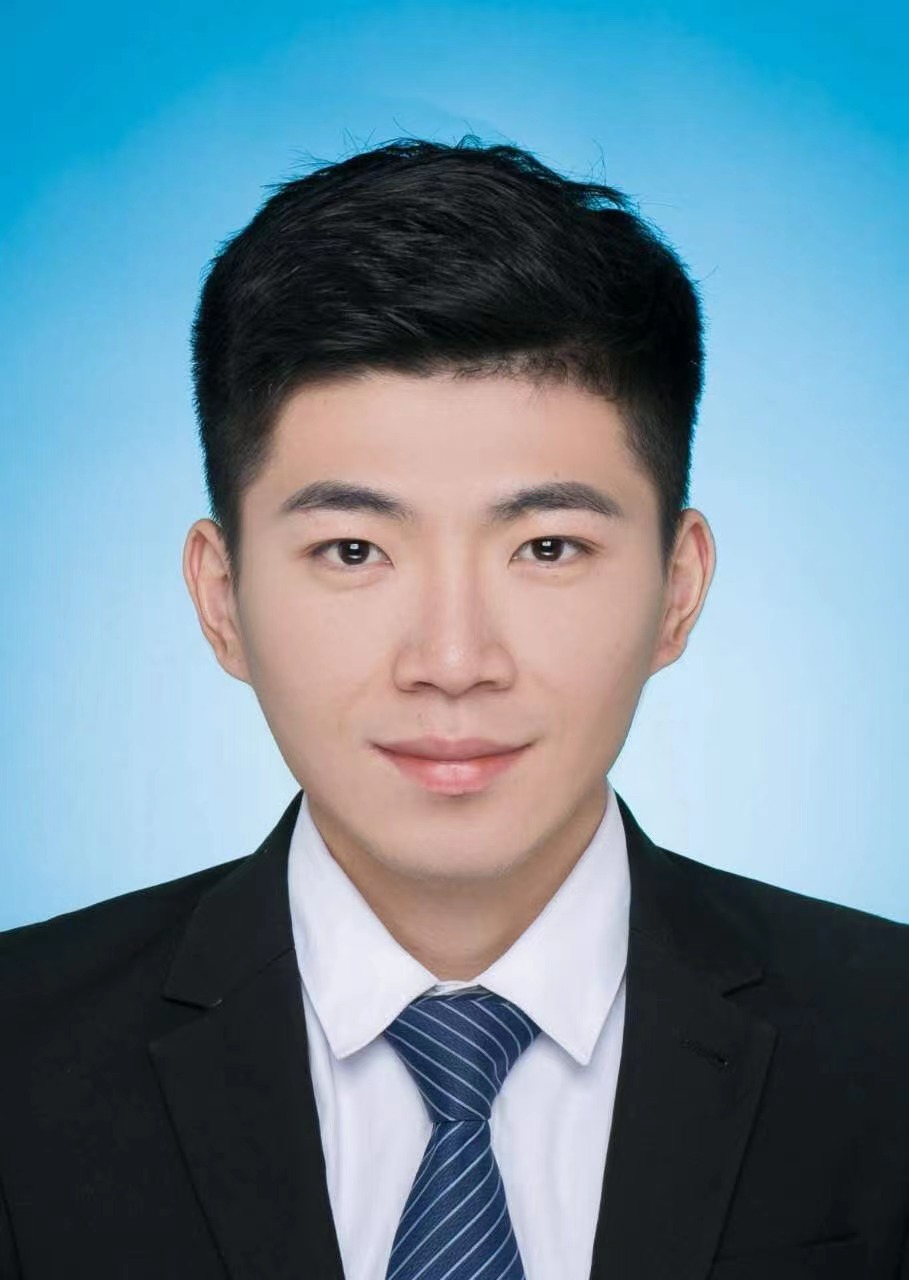}}]{Hanwei Zhu} received the B.E. and M.S. degrees from the Jiangxi University of Finance and Economics, Nanchang, China, in 2017 and 2020, respectively. He is currently pursuing a Ph.D. degree in the Department of Computer Science, City University of Hong Kong.  His research interests include perceptual image processing and computational photography.
\end{IEEEbiography}

\begin{IEEEbiography}[{\includegraphics[width=1in,height=1.25in,clip,keepaspectratio]{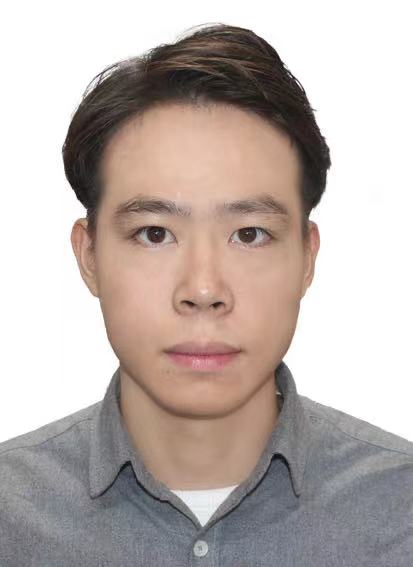}}]{Lingyu Zhu} received the B.Eng degree from the Wuhan University of Technology in 2018 and MA.Eng degree from Hong Kong University of Science and Technology in 2019. He is currently pursuing a Ph.D. degree at the City University of Hong Kong. His research interests include image/video quality assessment, image/video processing, and video compression.
\end{IEEEbiography}

\begin{IEEEbiography}[{\includegraphics[width=1in,height=1.25in,clip,keepaspectratio]{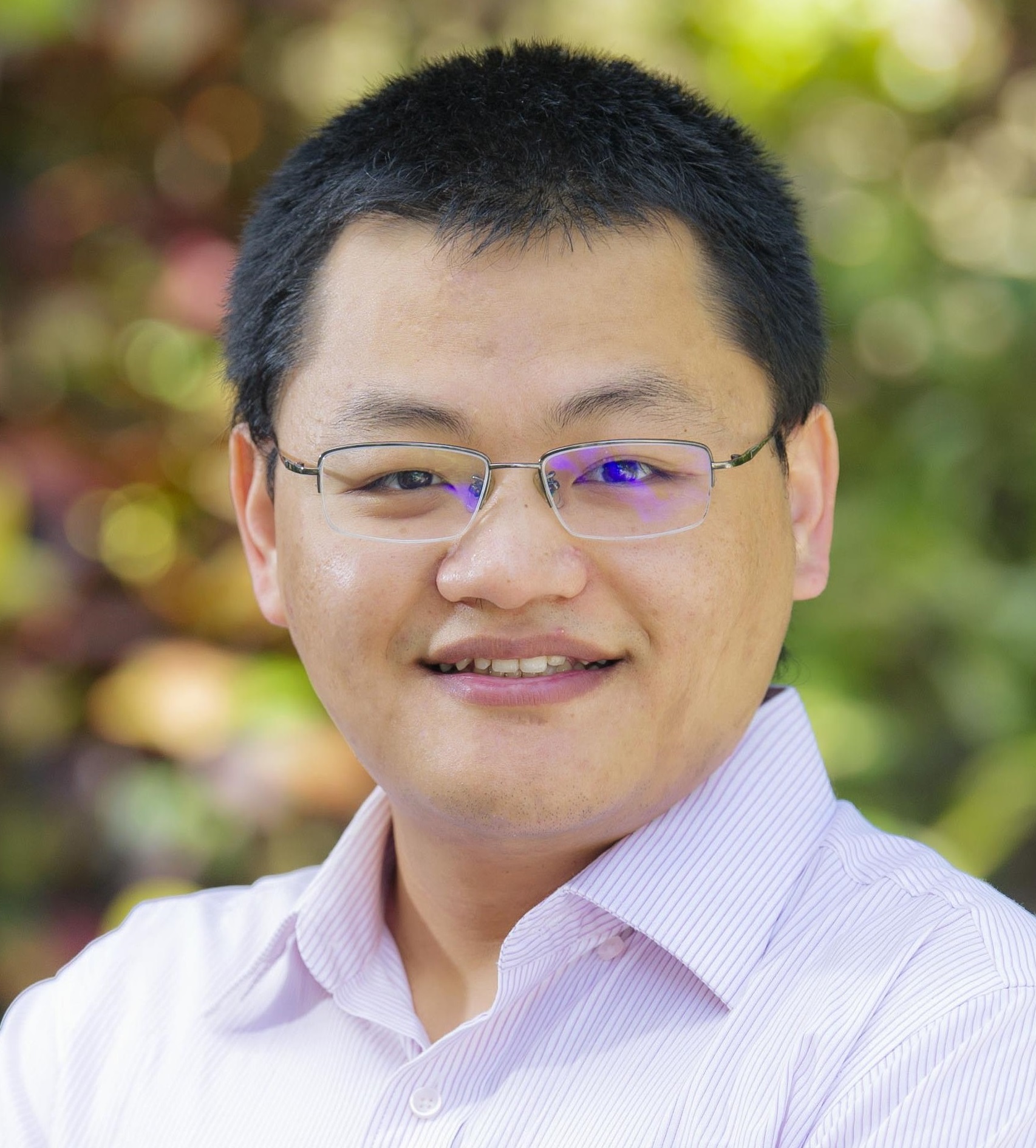}}]{Shiqi Wang} (Senior Member, IEEE) is currently an Associate Professor with the Department of Computer Science, City University of Hong Kong. He has proposed more than 50 technical proposals to ISO/MPEG, ITU-T, and AVS standards, and authored or coauthored more than 300 refereed journal articles/conference papers. His research interests include video compression, image/video quality assessment, and image/video search and analysis. He received the Best Paper Award from IEEE VCIP 2019, ICME 2019, IEEE Multimedia 2018, and PCM 2017. His co-authored article received the Best Student Paper Award in the IEEE ICIP 2018. He was a recipient of the 2021 IEEE Multimedia Rising Star Award in ICME 2021. 
\end{IEEEbiography}

\begin{IEEEbiography}[{\includegraphics[width=1in,height=1.25in,clip,keepaspectratio]{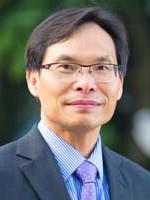}}]{Sam Kwong}
(Fellow, IEEE) received the B.S. degree in electrical engineering from the State University of New York, Buffalo, in 1983, the M.S. degree in electrical engineering from the University of Waterloo, Waterloo, ON, Canada, in 1985, and the Ph.D. degree from the University of Hagen, Germany, in 1996. From 1985 to 1987, he was a Diagnostic Engineer with Control Data Canada. He joined Bell Northern Research Canada as a member of Scientific Staff. In 1990, he became a Lecturer with the Department of Electronic Engineering, City University of Hong Kong, Hong Kong. Later, he was a Chair Professor at the Department of Computer Science, City University of Hong Kong. In 2023, he joined Lingnan University as the Chair Professor of Computational Intelligence and Associate Vice-President (Strategic Research). His research interests include video and image coding and evolutionary algorithms.
\end{IEEEbiography}
\end{document}